\documentclass[onecolumn]{article}         % twocolumn
\usepackage[top=3.6cm, bottom=3.2cm, left=2.5cm, right=2.5cm]{geometry}

\usepackage{color}
\usepackage{graphicx}
\usepackage{cite}
\usepackage[cmex10]{amsmath}
\usepackage{amssymb}
\usepackage{subfigure}
\usepackage{rotating}
\usepackage{multirow}
\usepackage{makecell}
\usepackage[colorlinks=true,linkcolor=red,citecolor=blue]{hyperref}
\begin{document}

\title{AID: A Benchmark Dataset for Performance Evaluation of Aerial Scene Classification}

\author{Gui-Song~Xia$^1$, Jingwen~Hu$^{1,2}$, Fan~Hu$^{1,2}$, Baoguang Shi$^3$, \\
Xiang~Bai$^3$, Yanfei~Zhong$^1$, Liangpei~Zhang$^1$\\
%\author{Gui-Song~Xia$^1$, Jingwen~Hu$^{1}$, Fan~Hu$^{1}$, Xiang~Bai$^2$\\
\\
$^1${\em State Key Lab. LIESMARS, Wuhan University, Wuhan, China.}\\
$^2${\em School of Electronic Information, Wuhan University, Wuhan, China.}\\
$^3${\em Electronic Information School, Huazhong University of Science and Technology, China.}
}

\maketitle
\begin{abstract}
Aerial scene classification, which aims to automatically label an aerial image with a specific semantic category, is a fundamental problem for understanding high-resolution remote sensing imagery. In recent years, it has become an active task in remote sensing area and numerous algorithms have been proposed for this task, including many machine learning and data-driven approaches. However, the existing datasets for aerial scene classification like UC-Merced dataset and WHU-RS19 are with relatively small sizes, and the results on them are already saturated. This largely limits the development of scene classification algorithms.
This paper describes the Aerial Image Dataset (AID): a large-scale dataset for aerial scene classification.
The goal of AID is to advance the state-of-the-arts in scene classification of remote sensing images. For creating AID, we collect and annotate more than ten thousands aerial scene images. In addition, a comprehensive review of the existing aerial scene classification techniques as well as recent widely-used deep learning methods is given. Finally, we provide a performance analysis of typical aerial scene classification and deep learning approaches on AID, which can be served as the baseline results on this benchmark.
\end{abstract}
%
%
%

%\vspace{-3mm}
\section{Introduction}
\label{sec:intro}
Nowadays, aerial images enable us to measure the earth surface with detail structures and are a kind of data source of great significance for earth observation~\cite{hu2013exploring,cheng2015effective,cheng2014multi,hu2015transferring}. Due to the drastically increasing number of aerial images and the highly complex geometrical structures and spatial patterns, to effectively understand the semantic content of them is particularly important, driven by many real-world applications in remote sensing community.  In this paper, we focus on aerial scene classification, a key problem in aerial image understanding, which aims to automatically assign a semantic label to each aerial image in order to know which category it belongs to.

The problem of aerial scene classification has received growing attention in recent years~\cite{risojevic2011aerial,yang2011spatial,sheng2012high,risojevic2012orientation,hu2013tile,luo2013indexing,
shao2013hierarchical,shao2013extreme,risojevic2013fusion,yang2013geographic,zhao2013scene,zhao2013hybrid,zheng2013automatic,
avramovic2014block,cheriyadat2014unsupervised,cheng2014multi,kusumaningrum2014integrated,negrel2014evaluation,zhao2014wavelet,
zhao2014land,zhu2014multi,cheng2015effective,chen2015pyramid,chen2015measuring,zhong2015scene,sridharan2015bag,hu2015comparative,
hu2015benchmark,hu2015transferring,hu2015unsupervised,castelluccio2015land,penatti2015deep,luus2015multiview,yang2015learning,zhang2015saliency,
zhang2015scene,zhu2015scene,chen2015land,zou2015deep,nogueira2016towards}.
In the literature, primary studies have devoted to classifying aerial images at pixel level, by assigning each pixel in an aerial image with a thematic class~\cite{tuia2009active,tuia2011survey}.
However, with the increasement of the spatial resolutions, it turns to be infeasible to interpret aerial images at pixel level~\cite{tuia2009active,tuia2011survey}, mainly due to the fact that single pixels quickly lose their thematic meanings and discriminative efficiency to separate different type of land covers.
Specifically, in 2001, Blaschke and Strobl~\cite{blaschke2001pixels} have raised the question ``{\em What's wrong with pixels?}" and argued that it is more efficient to analyze aerial images at object-level, where the ``objects" refer to local regions of pixels sharing spectral or texture homogeneities, e.g. superpixels~\cite{kotliar1990multiple}.
This kind of approaches then have dominated the analysis of high-resolution remote sensing images for decades~\cite{blaschke2003object,yan2006comparison,blaschke2010object,myint2011per,duro2012comparison,zhong2014hybrid,zhao2015detail}.
It is worth noticing that both pixel- and object-level classification methods attempt to model an aerial scene in a bottom-up manner by aggregating extracted spectral, texture and geometrical features for training a strong classifier.

However, due to the growing of image spatial resolutions, aerial scenes may often consist of different and distinct thematic classes~\cite{cheriyadat2014unsupervised} and it is of great interest to
reveal the context of these thematic classes, {\em i.e.} semantic information, of aerial scenes.
Aerial scene classification aims to classify an aerial image into different semantic categories by directly modeling the scenes by exploiting the variations in the spatial arrangements and structural patterns.
%by analyzing the patterns of different low-level visual features but regardless of the specific land cover types of each pixel or super-pixel.
In contrast with pixel-/object-oriented classification, scene classification provide a relatively high-level interpretation of aerial images.
More precisely, the item ``scene" hereby usually refers to a local area in large-scale aerial images that contain clear semantic information on the surface~\cite{yang2008comparing,dos2010evaluating,lienou2010semantic,xia2010structural,yang2010bag,chen2011evaluation,dai2011satellite,
risojevic2011gabor,risojevic2011aerial,yang2011spatial,sheng2012high,risojevic2012orientation,hu2013tile,luo2013indexing,
shao2013hierarchical,shao2013extreme,risojevic2013fusion,yang2013geographic,zhao2013scene,zhao2013hybrid,zheng2013automatic,
avramovic2014block,cheriyadat2014unsupervised,cheng2014multi,kusumaningrum2014integrated,negrel2014evaluation,zhao2014wavelet,
zhao2014land,zhu2014multi,cheng2015effective,chen2015pyramid,chen2015measuring,zhong2015scene,sridharan2015bag,hu2015comparative,
hu2015benchmark,hu2015transferring,hu2015unsupervised,castelluccio2015land,penatti2015deep,luus2015multiview,yang2015learning,zhang2015saliency,
zhang2015scene,zhu2015scene,chen2015land,zou2015deep,nogueira2016towards}. %, which consists of not only a single object like storage tanks, but also more complicated patterns like commercial area, residential area, etc.

%\subsection{Motivation and objective}

Though many exciting progresses on aerial scene classification have been extensivly reported in recent years, e.g. ~\cite{
avramovic2014block,cheriyadat2014unsupervised,cheng2014multi,kusumaningrum2014integrated,negrel2014evaluation,zhao2014wavelet,
zhao2014land,zhu2014multi,cheng2015effective,chen2015pyramid,chen2015measuring,zhong2015scene,sridharan2015bag,hu2015comparative,
hu2015benchmark,hu2015transferring,hu2015unsupervised,castelluccio2015land,penatti2015deep,luus2015multiview,yang2015learning,zhang2015saliency,
zhang2015scene,zhu2015scene,chen2015land,zou2015deep,nogueira2016towards}, there are two major issues that seriously limit the development of aerial scene classification.

\begin{itemize}
	\item[-] {\em Lacking a comprehensive review of existing methods.} Although many methods have been presented to advance the aerial scene classification, most of them were evaluated on different datasets under different experimental settings. This somewhat makes the progress confused and may misleads the development of the problem. Moreover, the codes of these algorithms have not been released, which brings difficulties to reproduce the works for fair comparisons. Therefore, the state-of-the-art of aerial scene classification is not absolutely clear.
	%Few work gives a comprehensively review and comparison concerning this exciting field, except~\cite{penatti2015deep,hu2015transferring,nogueira2016towards}, which provide partial reviews on popular deep-learning approaches for scene classification.
	\item[-] {\em Lacking proper benchmark datasets for performance evaluation.}
	In order to develop robust methods for aerial scene classification, it is highly expected that the datasets for evaluation demonstrate all the challenging aspects of the problem. For instance, the high diversities in the geometrical and spatial patterns of aerial scenes. Currently, the evaluations of aerial scene classification algorithms are typically done on datasets containing up to two-thousands images at best, e.g. the UC-Merced dataset~\cite{yang2010bag} and the WHU-RS19 dataset~\cite{xia2010structural}. Such limited number of images are critically insufficient to approximate the real applications, where the images are with high intra-class diversity and low inter-class variation. Recently, the saturated results on these datasets demonstrated that the more challenging datasets are badly required.
		
\end{itemize}

Due to the above issues,  in this paper, we present a comprehensive review to up-to-date algorithms as well as a new large-scale benchmark dataset of aerial images (named as AID), in order to fully advance the task of aerial scene classification. AID provides the research community a benchmark resource for the development of the state-of-the-art algorithms in aerial scene classification or other related tasks such as aerial image search, aerial image annotation, aerial image segmentation, etc. In addition, our experiments on AID demonstrate that it is quite helpful to reflect the shortcomings of existing methods. In summary, the major contributions of this paper are as follows:

\begin{itemize}
  \item[-] We provide a comprehensive review on aerial scene classification, by giving a clear summary of the development of scene classification approaches. %and providing source codes for their implementations.
  \item[-] We construct a new large-scale dataset, {\em i.e.} AID, for aerial scene classification. The dataset is, to our knowledge, of the largest size and the images are with high intra-class diversity and low inter-class dissimilarity, which can provide the research community a better data resource to evaluate and advance the state-of-the-art algorithms in aerial image analysis.
  \item[-] We evaluate a set of representative aerial scene classification approaches with various experimental protocols on the new dataset. These can serve as baseline results for future works.
  \item[-] The source codes of our implementation for all the baseline algorithms are released, will be general tools for other researchers.
\end{itemize}

The rest of the paper is organized as follows. We first provide a comprehensive review of related methods in Section~\ref{sec:review}. The details of AID dataset are described in Section~\ref{sec:aid}. Then, we provide the description of baseline algorithms for benchmark evaluation in Section~\ref{sec:baseline}. In Section~\ref{sec:expertment}, the evaluation and comparison of baseline algorithms on AID under different experimental settings are given. Finally, some conclusion remarks are drawn in Section~\ref{sec:conclusion}.

The AID and the codes for reproducing all the results in this paper are downloadable at the project webpage~\url{www.lmars.whu.edu.cn/xia/AID-project.html}.

%%%
%%%

%\vspace{-2mm}
\section{A review on aerial scene classification}
\label{sec:review}
This section reviews comprehensively the existing scene classification methods for aerial images.
Distinguished from pixel-/object-level image classifications which interpret aerial images with a bottom-up manner, scene classification is apt to directly model an aerial scene by developing a holistic representation of the aerial image~\cite{yang2008comparing,dos2010evaluating,lienou2010semantic,xia2010structural,yang2010bag,chen2011evaluation,dai2011satellite,
risojevic2011gabor,risojevic2011aerial,yang2011spatial,sheng2012high,risojevic2012orientation,hu2013tile,luo2013indexing,
shao2013hierarchical,shao2013extreme,risojevic2013fusion,yang2013geographic,zhao2013scene,zhao2013hybrid,zheng2013automatic,
avramovic2014block,cheriyadat2014unsupervised,cheng2014multi,kusumaningrum2014integrated,negrel2014evaluation,zhao2014wavelet,
zhao2014land,zhu2014multi,cheng2015effective,chen2015pyramid,chen2015measuring,zhong2015scene,sridharan2015bag,hu2015comparative,
hu2015benchmark,hu2015transferring,hu2015unsupervised,castelluccio2015land,penatti2015deep,luus2015multiview,yang2015learning,zhang2015saliency,
zhang2015scene,zhu2015scene,chen2015land,zou2015deep,nogueira2016towards}. One should observe that, actually, the underlying assumption of scene classification is that the same type of scene should share certain statistically holistic visual characteristics. This point has been verified on natural scenes~\cite{oliva2001modeling} and demonstrates its efficiency on classifying aerial scenes. Thus, most of the works on aerial scene classification focus on computing such holistic and statistical visual attributes for classification features.
In this sense, scene classification methods can be divided into three main categories: methods using low-level visual features, methods relying on mid-level visual representations and the methods based on high-level vision information. In what follows, we review each category of the methods in details.

\subsection{Methods using low-level visual features}
With this kind of methods, it is supposed that aerial scenes can be distinguished by low-level visual features, e.g. spectral, texture and structure, etc.
Consequently, an aerial scene image is usually described by a feature vector extracted from such low-level visual attributes, either locally~\cite{yang2008comparing} or globally~\cite{dos2010evaluating}.
On one hand, in order to describe the complex structures, local structure descriptors, e.g. the {\em Scale Invariant Feature Transform} (SIFT)~\cite{lowe2004sift}, have been widely used for modeling the local variations of structures in aerial images.
The classification feature vector is usually formed by concatenating or pooling the local descriptors of the subregions of an image.
On the other hand, for depicting the spatial arrangements of aerial scenes, statistical and global distributions of certain spatial cues such as color~\cite{swain1991color} and texture information~\cite{manjunath1996texture,risojevic2011aerial} have also been well investigated.
For instance, Yang and Newsam~\cite{yang2008comparing} compared SIFT and Gabor texture features for classifying IKONOS satellite images by using Maximum A Posteriori (MAP) classifier and found that SIFT performs better. Santos {\em et. al}~\cite{dos2010evaluating} evaluated various global color descriptors and texture descriptors, e.g. color histogram~\cite{swain1991color}, local binary pattern (LBP)~\cite{ojala2002lbp}, for scene classification.

Although single type of features work well for aerial image classification~\cite{yang2008comparing,dos2010evaluating,risojevic2011gabor}, the combinations of complementary features can often improve the results, see e.g.~\cite{luo2013indexing,avramovic2014block,chen2015measuring}. In particular, Luo {\em et. al}~\cite{luo2013indexing} extracted $6$ different kinds of feature descriptors, {\em i.e.} simple radiometric features, Gaussian wavelet features~\cite{luo2008indexing}, Gray Level Co-Occurrence Matrix (GLCM), Gabor filters, shape features~\cite{luo2009local} and SIFT, and combined them to form a multiple-feature representation for indexing remote sensing images with different spatial resolutions, which reported that multiple features can describe aerial scenes better. Avramovic and Risojevic integrated Gist~\cite{oliva2001modeling} and SIFT descriptors for aerial scene classification.
Some class-specific feature selection methods were also developed to select a good subset of low-level visual features for aerial image classification~\cite{chen2015measuring}.

In order to encode the global spatial arrangements and the geometrical diversities of aerial scenes, Xia {\em et. al}~\cite{xia2010structural} proposed an invariant and robust shape-based scene descriptor to describe the structure distributions of aerial images. While Vladimir {\em et. al} focused on the texture information of scenes and they successively proposed a local structural texture descriptor~\cite{risojevic2011aerial}, an orientation difference descriptor~\cite{risojevic2012orientation}, and an Enhanced Gabor Texture Descriptor (EGTD)~\cite{risojevic2013fusion} based on the Gabor filters~\cite{manjunath1996texture} to further improve the performance. In~\cite{chen2015land}, a multi-scale completed local binary patterns (MS-CLBP) was proposed for land-use scene classification and achieved the state-of-the-art performance among low-level methods.

It is worth noticing that scene classification methods with low-level visual features perform well on some aerial scenes with uniform structures and spatial arrangements, but it is difficult for them to depict the high-diversity and the non-homogeneous spatial distributions in aerial scenes~\cite{yang2013geographic}.

\subsection{Methods relying on mid-level visual representations}

In contrast with methods relying on low-level visual attributes, mid-level aerial scene analysis approaches
attempt to develop a holistic scene representation through representing the high-order statistical patterns formed by the extracted local visual attributes.
A general pipeline is to first extract local image attributes, e.g. SIFT, LBP and color histograms of local image patches, and then encode these local cues for building a holistic mid-level representation for aerial scenes.

One of the most popular mid-level approaches is the {\em bag-of-visual-words} (BoVW) model~\cite{yang2010bag}.
More precisely, this method~\cite{yang2010bag} first described local image patches by SIFT~\cite{lowe2004sift} descriptors, then learned a vocabulary of visual words (also known as dictionary or codebook) for instance by k-means clustering. Subsequently, the local descriptors were encoded against the vocabulary by hard assignment, {\em i.e.}, vector quantization, and a global feature vector of the image could be obtained by the histogram of visual words, which is actually counting the occurrence frequencies of each visual words in the image.
Thanks to its simplicity and efficiency, BoVW model and its variants have been widely adopted for computing mid-level representation for aerial scenes, see e.g.~\cite{yang2010bag,yang2011spatial,chen2011evaluation,shao2013hierarchical,shao2013extreme,negrel2014evaluation,
zhao2014land,zhao2014wavelet,chen2015pyramid,hu2015benchmark,sridharan2015bag}.

In order to improve the discriminative power of BoVW model, multiple complemented low-level visual features were combined under the framework. For instance, in~\cite{chen2011evaluation} various local descriptors, including SIFT, GIST, color histogram and LBP, {\em etc.}, were evaluated with the standard BoVW model for aerial scene classification. The experiments of the concatenation of BoVW representations from different local descriptors proved that combining complemented features can significantly improve the classification accuracy.
Similarly, in~\cite{shao2013extreme}, a highly discriminative texture descriptor, {\em i.e.} combined scattering feature~\cite{mallat2012combined}, was incorporated with SIFT and color histogram to extract structure and spectral information under a multi-feature extraction scheme. Unlike the simple concatenation in~\cite{chen2011evaluation}, in this work, a hierarchical classification method incorporating {\em Extreme Value Theory} (EVT)-based normalization~\cite{scheirer2012evt} was used to calibrate multiple features.
In~\cite{sheng2012high}, multiple features like structure features, spectral features and texture features were extracted and encoded in a sparse coding scheme~\cite{yang2009scspm} besides BoVW. The sparse coding scheme was developed based on BoVW but adding a sparsity constraint to the feature distributions to reduce the complexity of {\em Support Vector Machine} (SVM) meanwhile maintain good performance. In~\cite{negrel2014evaluation}, various feature coding methods developed from BoVW model were evaluated for scene classification using multi-features. By applying PCA for dimension reduction before concatenating multi-features, the {\em Improved Fisher Vector} (IFK)~\cite{Perronnin2010ifk} and {\em Vectors of Locally Aggregated Tensors} (VLAT)~\cite{negrel2014evaluation} methods reported to achieve the state-of-the-art performances.

Note that in BoVW models, they count the frequencies of the visual words in an image, with regardless of the spatial distribution of the visual words. However, the spatial arrangements of visual words, e.g. co-occurrence, convey important information of aerial scenes.
Therefore, some methods were proposed to incorporate the spatial distribution of visual words beyond the BoVW models. For instance, Yang {\em et al}~\cite{yang2011spatial} developed the {\em spatial pyramid co-occurrence kernel} (SPCK) to integrate the absolute and relative spatial information ignored in the standard BoVW model setting, by relying on the idea of {\em spatial pyramid match kernel} (SPM)~\cite{lazebnik2006spm} and {\em spatial co-occurrence kernel} (SCK)~\cite{yang2010bag}.
Later, Zhao {\em et. al}~\cite{zhao2014wavelet} proposed another way to incorporate the spatial information, where wavelet decomposition was utilized in the BoVW model to combine not only the spatial information but also the texture information.
Moreover, in~\cite{zhao2014land}, the authors proposed a concentric circle-based spatial-rotation-invariant representation to encode the spatial information. In~\cite{chen2015pyramid}, a {\em pyramid-of-spatial-relatons} (PSR) model was developed to capture both absolute and relative spatial relationships of local low-level features. Unlike the conventional co-occurrence approaches~\cite{yang2011spatial,zhao2014land} that describe pairwise spatial relationships between local features, the PSR model employed a novel concept of spatial relation to describe relative spatial relationship between a group of local features and reported better performance.

In addition, to encode higher-order spatial information between low-level local visual words for scene modeling, topic models along with the BoVW scheme are developed to take into account the semantic relationship among the visual words~\cite{lienou2010semantic,hu2013tile,zhao2013hybrid,kusumaningrum2014integrated,zhong2015scene,zhu2015scene}.
Among them, {\em Latent Dirichlet Allocation} (LDA)~\cite{blei2003lda} model defines an intermediate variable named ``topic", which serves as a connection between the visual words and the image. The probability distribution of the topics are estimated by Dirichlet distribution and were used to describe an image instead of the marginal distribution of visual words with much lower dimensional features.
To combine different features, Kusumaningrum {\em et. al}~\cite{kusumaningrum2014integrated} used CIELab color moments~\cite{stricker1995similarity}, GLCM~\cite{haralick1973textural} and {\em edge orientation histogram} (EOH)~\cite{kusumaningrum2014integrated} to extract spectral, texture and structure information respectively in the LDA model.
%In particular, {\em Gaussian Mixture Model} (GMM)~\cite{stauffer1999gmm}, instead of the traditional k-means clustering algorithm and the specific discriminative visual words for each class were obtained by building an integrated visual vocabulary instead of the common used universal one, both can help to increase the classification performance.
In~\cite{zhong2015scene}, the {\em probabilistic Latent Semantic Analysis} (pLSA) model~\cite{bosch2006plsa} was adopted for scene classification in a multi-feature fusion manner and achieved better result than single feature. In~\cite{zhu2015scene}, pLSA~\cite{bosch2006plsa} and LDA~\cite{blei2003lda} were compared using a multi-feature fusion strategy to combine three complementary features in a semantic allocation level, and LDA demonstrated slightly better performances.

Observe that, in all the aforementioned methods, various hand-craft local image descriptors are used to represent aerial scenes. One main difficulty of such methods lies in the fact that they may lack the flexibility and adaptivity to different scenes. In this sense, unsupervised feature learning approaches have been developed to automatically learn adaptive feature representations from images, see e.g.~\cite{cheriyadat2014unsupervised,hu2015unsupervised,zhang2015saliency}. In~\cite{cheriyadat2014unsupervised}, a sparse coding based method was proposed to learn a holistic scene representation from raw pixel values and other low-level features for aerial images. In~\cite{hu2015unsupervised}, Hu {\em et. al} discovered the intrinsic space of local image patches by applying different manifold learning techniques and made the dictionary learning and feature encoding more effective.
Also with the unsupervised feature learning scheme, Zhang {\em et. al}~\cite{zhang2015saliency} extracted the features of image patches by the {\em sparse Auto-Encoder} (SAE)~\cite{vincent2010ae} and exploited the local spatial and structural information of complex aerial scenes.

\subsection{Methods based on high-level vision information}

Currently, deep learning methods achieve impressive results on many computer vision tasks such as image classification, object and scene recognition, image retrieval, etc.
This type of methods also achieve state-of-the-art performance on aerial scene classification, see e.g.~\cite{hu2015transferring,castelluccio2015land,penatti2015deep,luus2015multiview,zhang2015scene,zou2015deep,nogueira2016towards}.
In general, deep learning methods use a multi-stage global feature learning architecture to adaptively learn image features and often cast the aerial scene classification as an end-to-end problem. Compared with low-level and mid-level methods, deep learning methods can learn more abstract and discriminative semantic features and achieve far better classification performance~\cite{hu2015transferring,castelluccio2015land,penatti2015deep,luus2015multiview,zhang2015scene,nogueira2016towards}.

It has been reported that by directly using the pre-trained deep neural network architectures on the natural images~\cite{ILSVRC15}, the extracted global features showed impressive performance on aerial scene classification~\cite{penatti2015deep}. The two freely available pre-trained deep Convolution Neural Network (CNN) architectures are OverFeat~\cite{sermanet2013overfeat} and CaffeNet~\cite{jia2014caffe}.
In~\cite{castelluccio2015land}, another promising architecture, {\em i.e.} GoogLeNet~\cite{szegedy2014going}, was considered and evaluated. This architecture also showed astounding performance for aerial images. In~\cite{luus2015multiview}, it demonstrated that a multi-scale input strategy for multi-view deep learning can improve the performance of aerial scene classification.

In contrast with directly using the features from the fully-connected layer of the pre-trained CNN architectures as the final representation~\cite{castelluccio2015land,penatti2015deep,luus2015multiview},
others use the deep-CNN as local feature extractor and combine it with feature coding techniques. For instance, Hu {\em et. al}~\cite{hu2015transferring} extracted multi-scale dense CNN activations from the last convolutional layer as local features descriptors and further coded them using feature encoding methods like BoVW~\cite{sivic2003bow}, {\em Vector of Locally Aggregated Descriptors} (VLAD)~\cite{jegou2012vlad} and {\em Improved Fisher Kernel} (IFK)~\cite{Perronnin2010ifk} to generate the final image representation. For all the deep-CNN architectures used above, either the global or local features were obtained from the networks pre-trained on natural image datasets and were directly used for classification aerial images.
%Both can achieve the state-of-the-art results, which showes the high generalization ability of the deep learning methods.

In addition to the above two ways using deep-learning methods, another choice is to train a new deep network. However, as reported in~\cite{nogueira2016towards}, using the existing aerial scene datasets (e.g. UC-Merced dataset~\cite{yang2010bag} and the WHU-RS19 dataset~\cite{xia2010structural}) to fully train the networks such as CaffeNet~\cite{jia2014caffe} or GoogLeNet~\cite{szegedy2014going} showed a drop in accuracies compared with using the networks as global feature extractors.
This can be explained by the fact that the large scale networks usually contain millions of parameters to be trained, therefore, to train them using the aerial datasets with only a few hundreds or thousands images will easily stick in overfitting and local minimum.
Thus, to better fit the dataset, smaller networks for classification were trained~\cite{zhang2015scene,zou2015deep}. In~\cite{zhang2015scene}, a {\em Gradient Boosting Random Convolutional Network} (GBRCN) was proposed for classifying aerial images with only two convolutional layers. In~\cite{zou2015deep}, a {\em deep belief network} (DBN)~\cite{Hinton2006A} was trained on aerial images, and the feature selection problem was formulated as a feature reconstruction problem in the DBN scheme. By minimizing the reconstruction error over the whole feature set, the features with smaller reconstruction errors can hold more feature intrinsics for image representation.
%However, to train a new deep network is very time-consuming, e.g., to train a two layer CNN usually needs several hours and even several days, the result of which is just comparable with the networks pre-trained.
However, the generalization ability of a small network is often lower than that of large scale networks. It is highly demanded to train a large-scale network with large number of annotated aerial images.

%%%%%%%%
%%%%%%%%
\section{Aerial Image Datasets (AID) for \\ aerial scene classification}
\label{sec:aid}
This section first reviews several datasets commonly used for aerial scene classification and then described the proposed Aerial Image Dataset (AID)\footnote{The AID is downloadable at~\url{www.lmars.whu.edu.cn/xia/AID-project.html}.}.

%\begin{figure*}[htb!]
%  \centering
%  \includegraphics[width= 0.8\linewidth]{figure/ucm-samples.pdf}
%  \caption{Samples of UC-Merced dataset: three examples of each semantic scene class are shown. There are 21 classes with 100 samples per class.}
%  \label{ucm-samples}
%\end{figure*}

\subsection{Existing datasets for aerial scene classification}
%The number of public datasets is considerably small.

\subsubsection{UC-Merced dataset~\cite{yang2010bag}} It consists of 21 classes of land-use images selected from aerial ortho-imagery with the pixel resolution of one foot. %Three samples of each class are shown in Fig.~\ref{ucm-samples}.
The original images were downloaded from the United States Geological Survey (USGS) National Map of the following US regions: {\em Birmingham, Boston, Buffalo, Columbus, Dallas, Harrisburg, Houston, Jacksonville, Las Vegas, Los Angeles, Miami, Napa, New York, Reno, San Diego, Santa Barbara, Seattle, Tampa, Tucson, and Ventura}. They are then cropped into small regions of $256 \times 256$ pixels. There are totally $2100$ images manually selected and uniformly labeled into $21$ classes: {\em agricultural, airplane, baseball diamond, beach, buildings, chaparral, dense residential, forest, freeway, golf course, harbor, intersection, medium density residential, mobile home park, overpass, parking lot, river, runway, sparse residential, storage tanks, and tennis courts}.

It is worth noticing that
UCM dataset contains a variety of spatial land-use patterns which make the dataset more challenging.
Moreover, some highly overlapped classes, e.g. dense residential, medium residential and sparse residential that mainly differ in the density of structures, make the dataset difficult for classification.
This dataset is widely used for the task of aerial image classification~\cite{yang2010bag,risojevic2011aerial,yang2011spatial,risojevic2012orientation,
shao2013hierarchical,risojevic2013fusion,yang2013geographic,zheng2013automatic,
avramovic2014block,cheriyadat2014unsupervised,cheng2014multi,negrel2014evaluation,zhao2014wavelet,
zhao2014land,zhu2014multi,cheng2015effective,chen2015pyramid,chen2015measuring,hu2015comparative,
hu2015transferring,hu2015unsupervised,castelluccio2015land,penatti2015deep,luus2015multiview,yang2015learning,zhang2015saliency,
zhang2015scene,zhu2015scene}.

%%Thus, the main advantages of this dataset lies in the fact that it is the largest and most challenging public dataset for HSR-RS image scene classification. However, the main defects lie in the fact that all the sample images are from US regions with a fixed resolution, thus, it may result in the problem that the methods perform well on this dataset may be fitted to the dataset and cannot be generalized to other countries or different resolutions.

%\begin{figure*}[htb!]
%  \centering
%  \includegraphics[width= .8\linewidth]{figure/rs19-samples.pdf}
%  \caption{Samples of WHU-RS dataset (the second version): three examples of each semantic scene class are shown. There are 19 classes with 50 samples per class.}
%  \label{rs19-samples}
%\end{figure*}

\subsubsection{WHU-RS dataset} This dataset is collected from Google Earth imagery\footnote[3]{https://www.google.com/earth/}. The images are with fixed size of $600 \times 600$ pixels with various pixel resolutions up to half a meter.
This dataset has been updated to the third versions until now.
%Some samples per each class of the second version are displayed in Fig.~\ref{rs19-samples}.
In its original version~\cite{xia2010structural}, there are $12$ classes of aerial scenes including {\em airport, bridge, river, forest, meadow, pond, parking, port, viaduct, residential area, industrial area}, and {\em commercial area}. For each class, there were $50$ samples. Later, Sheng {\em et.al}~\cite{sheng2012high} expanded the dataset to $19$ classes with $7$ new ones, i.e. {\em beach, desert, farmland, football field, mountain, park} and {\em railway station}. Thus, the dataset is composed of a total number of $950$ aerial images, which is widely used as the {\em WHU-RS19 dataset}~\cite{chen2011evaluation,sheng2012high,shao2013hierarchical,
hu2015comparative,hu2015benchmark,hu2015transferring,hu2015unsupervised,yang2015learning}.
However, the size of this dataset is relatively small compared with UC-Merced dataset~\cite{yang2010bag}. Thus, we reorganized and expanded WHU-RS19 to form its third version~\cite{hu2015benchmark}, by adding a new aerial scene type ``bare land" and increasing the number of samples in each class.
In the newest version of the WHU-RS dataset, it thus has $5000$ aerial images with each class containing more than $200$ sample images.
It is worth noticing that the sample images of the same class in WHU-RS dataset are collected from different regions all around the world and the aerial scenes might appear at different scales, orientations and with different lighting conditions.
%Therefore, this dataset can make up the defects of the UC-Merced dataset. Nonetheless, the inter-class distance increases for it lacks the highly overlapping scene classes as in the UC-Merced dataset, thus, it is less challenging compared to the UC-Merced dataset.

\subsubsection{RSSCN7 dataset~\cite{zou2015deep}}This dataset is also collected from Google Earth\footnote[3]{https://www.google.com/earth/}, which contains $2800$ aerial scene images labeled into $7$ typical scene categories, i.e., the grassland, forest, farmland, parking lot, residential region, industrial region, river and lake. There are $400$ images in each scene type, and each image has a size of $400 \times 400$ pixels. It is worth noticing that the sample images in each class are sampled on $4$ different scales with $100$ images per scale with different imaging angles, which is the main challenge of the dataset.

\subsubsection{Other small datasets} Besides the three public datasets mentioned before, there are also several non-public datasets, e.g., the {\em IKONOS Satellite Image dataset}~\cite{yang2008comparing}, the In-House dataset~\cite{risojevic2011gabor,risojevic2011aerial,avramovic2014block}, the SPOT Image dataset~\cite{luo2013indexing}, the ORNL dataset~\cite{cheriyadat2014unsupervised}, etc. Note that the numbers of scene types in all these datasets are less than $10$, which thus results in small intra-class diversity. Moreover, common used mid-level scene classification methods get saturated and have reported overall accuracies nearly $100\%$ on these dataset. Therefore, these less challenging datasets will severely restrict the development of aerial scene classification algorithms.

\subsection{AID: a new dataset for aerial scene classification}
\begin{figure*}[htb!]
  \centering
  \includegraphics[width= 0.88\linewidth]{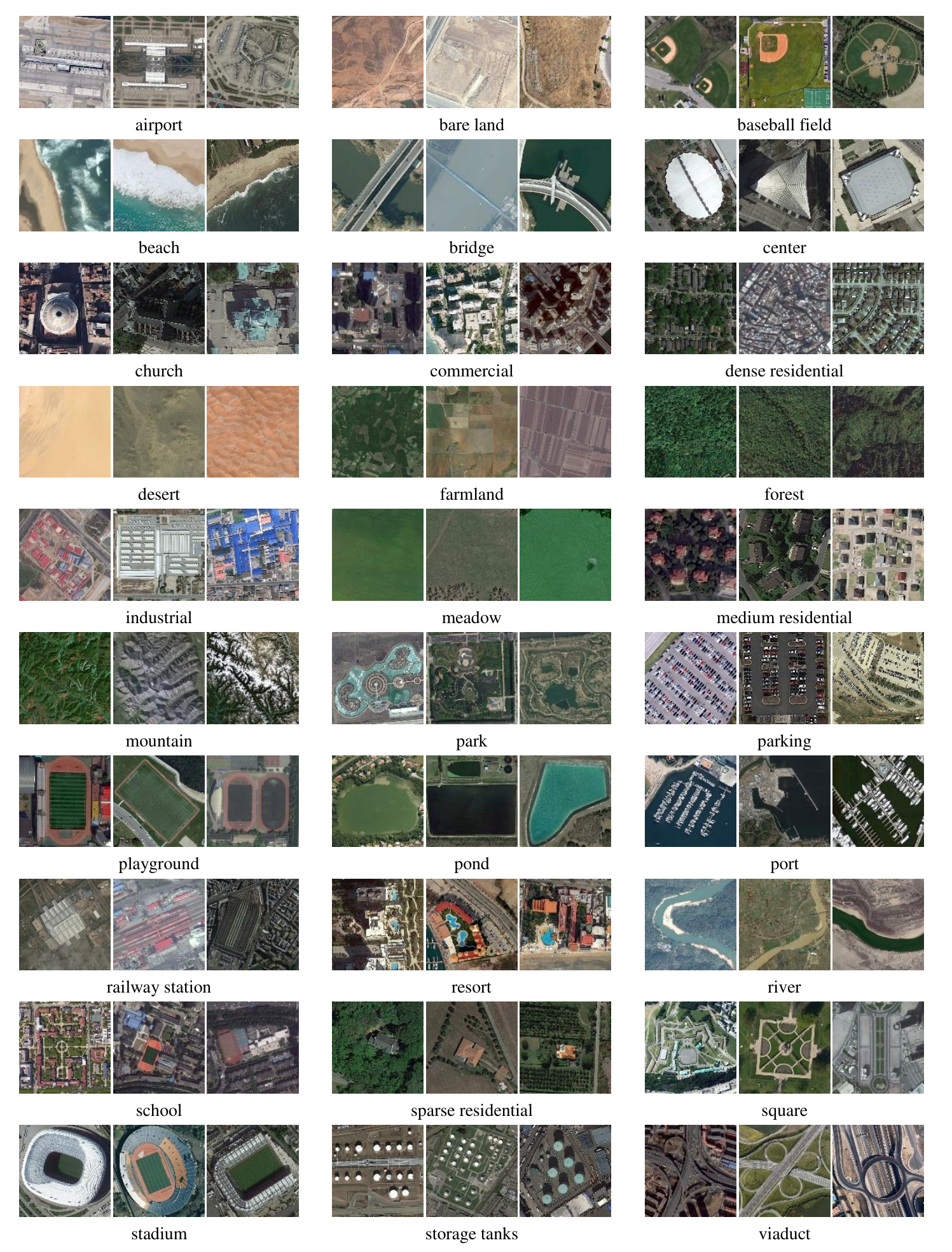}
  \caption{Samples of AID: three examples of each semantic scene class are shown. There are $10000$ images within $30$ classes.}
  \label{whu-samples}
\end{figure*}

%Owing to the high necessity of constructing new challenging dataset, we construct a new and large scale dataset to evaluate the performances of the scene classification algorithms more fairly, add more challenges and aim to promote the development of new methods.
To advance the state-of-the-arts in scene classification of remote sensing images, we construct AID, a new large-scale aerial image dataset, by collecting sample images from Google Earth imagery.
%It is known that Google Earth images are from both satellite images and aerial ones.
%For its satellite images, some come from a commercial satellite - QuickBird, which is launched by an United States company - DigitalGlobe; and some come from the EarthSat corporation, which is contracted with NASA's Landsat program. While the sources of the aerial images are from the BlueSky company, Sanborn company, etc.
Note that although the Google Earth images are post-processed using RGB renderings from the original optical aerial images, Hu {\em et al.}~\cite{hu2013exploring} have proven that there is no significant difference between the Google Earth images with the real optical aerial images even in the pixel-level land use/cover mapping. Thus, the Google Earth images can also be used as aerial images for evaluating scene classification algorithms.

The new dataset is made up of the following $30$ aerial scene types: {\em airport, bare land, baseball field, beach, bridge, center, church, commercial, dense residential, desert, farmland, forest, industrial, meadow, medium residential, mountain, park, parking, playground, pond, port, railway station, resort, river, school, sparse residential, square, stadium, storage tanks and viaduct}. All the images are labelled by the specialists in the field of remote sensing image interpretation, and some samples of each class are shown in Fig.~\ref{whu-samples}.
The numbers of sample images varies a lot with different aerial scene types, see Table.~\ref{tab:class}, from $220$ up to $420$. In all, the AID dataset has a number of $10000$ images within $30$ classes.

The images in AID are actually multi-source, as Google Earth images are from different remote imaging sensors. This brings more challenges for scene classification than the single source images like UC-Merced dataset~\cite{yang2010bag}.
Moreover, all the sample images per each class in AID are carefully chosen from different countries and regions around the world, mainly in {\em China, the United States, England, France, Italy, Japan, Germany}, etc., and they are extracted at different time and seasons under different imaging conditions, which increases the intra-class diversities of the data.
%To decrease the inter-class distance, we further add ten types of challenging scenes in the dataset, i.e., baseball field, center, church, medium residential, playground, resort, school, sparse residential, square, storage tanks.

Note that another main difference between AID and UC-Merced dataset is that AID has multi-resolutions: the pixel-resolution changes from about $8$ meters to about half a meter, and thus the size of each aerial image is fixed to be $600 \times 600$ pixels to cover a scene with various resolutions.

\begin{table*}[htb!]
\caption{The different semantic scene classes and the number of images in each class of the new dataset.}
\label{tab:class}
\centering
\vspace{1.5mm}
\begin{tabular}{c c|c c|c c}
\hline
  Types             & \#images  & Types              & \#images & Types              & \#images \\ \hline
  airport           & 360       & farmland           & 370      & port               & 380      \\
  bare land         & 310       & forest             & 250      & railway station    & 260      \\
  baseball field    & 220       & industrial         & 390      & resort             & 290      \\
  beach             & 400       & meadow             & 280      & river              & 410      \\
  bridge            & 360       & medium residential & 290      & school             & 300      \\
  center            & 260       & mountain           & 340      & sparse residential & 300      \\
  church            & 240       & park               & 350      & square             & 330      \\
  commercial        & 350       & parking            & 390      & stadium            & 290      \\
  dense residential & 410       & playground         & 370      & storage tanks      & 360      \\
  desert            & 300       & pond               & 420      & viaduct            & 420      \\
\hline
\end{tabular}
\end{table*}

\subsection{Why AID is proper for aerial image classification?}
In contrast with existing remote sensing image datasets, e.g. UC-Merced dataset and WHU-RS19 dataset, AID has following properties:

\begin{itemize}
  \item [-]\textbf{\emph{Higher intra-class variations:}}
In aerial images, due to the high spatial resolutions, the geometrical structures of scenes become more clear and bring more challenges to image classification. First, thanks to the high complexity of the earth surface, objects in the same type of scene may appear at different sizes and orientations.
Second, the different imaging conditions, e.g. the flying altitude and direction and the solar elevation angles, may also vary a lot the appearance of the scene. Thus, in order to develop robust aerial image classification algorithms with stronger generalized capability, it is accepted that the dataset contains high intra-class diversity.
  The increasing numbers of sample images per each class in AID allow us to collect images from different regions all over the world accompanied with different scales, orientations and imaging conditions, which can increase the intra-class diversities of the dataset, see e.g. Fig.~\ref{intra}.
  In Fig.~\ref{intra}.(a), we illustrate two examples of the same scene with different scale.   In Fig.~\ref{intra}.(b), we display examples of the same type of scene with different building styles, as the sample images are collected in different regions and countries and the appearances of the same scene varies a lot due to the cultural differences.
  In Fig.~\ref{intra}.(c), the shadow direction of the buildings vary from west to north at different imaging time; and a mountain varies from green to white along with the seasonal variation.

  \begin{figure}[htb!]
  \centering
  \includegraphics[width= .9  \linewidth]{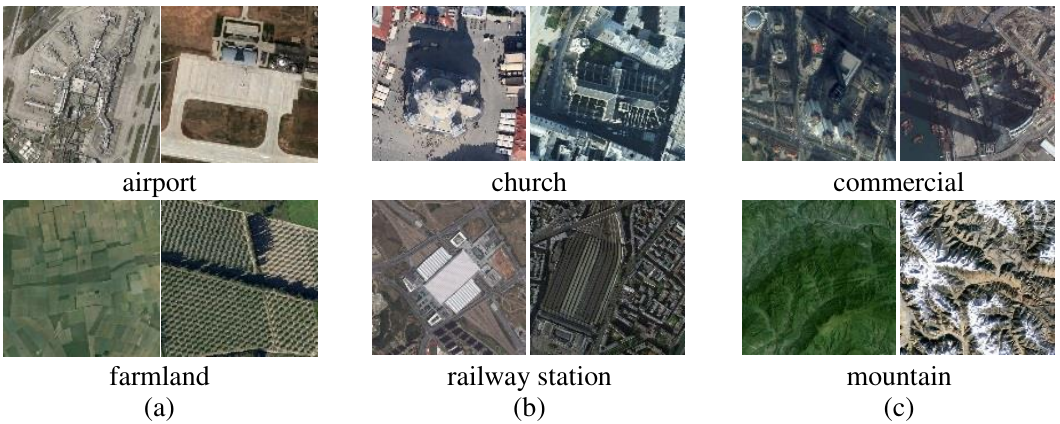}
  \caption{Large intra-class diversity:(a). multi-scale images of the same scene; (b). different building styles of the same scene; (c). different imaging conditions of the same scene.}
  \label{intra}
  \end{figure}

  \item [-]\textbf{\emph{Smaller inter-class dissimilarity:}} In real cases of aerial image classification, the dissimilarities between different scene classes are often small. The construction of AID well considered this point, by adding more scene categories.
As displayed in Fig.~\ref{inter}, AID contains scenes sharing similar objects. e.g., both {\em stadium} and {\em playground} may contain sports field (see Fig.~\ref{inter}.(a)), but the main difference lies in whether there are stands around.
Both {\em bare land} and {\em desert} are fine-grained textures and share similar colors (see Fig.~\ref{inter}.(b)), but {\em bare land} usually has more artificial traces. Some scene classes have similar structural distributions, like {\em resort} and {\em park} (see Fig.~\ref{inter}.(c)), which may contain a lake and some buildings, etc., however, a park is generally equipped with amusement and leisure facilities while a resort is usually composed of villas for vacations.
The AID has taken into account many these kinds of scene classes with small inter-class dissimilarity and makes it closer to real aerial image classification tasks.

  \begin{figure}[htb!]
  \centering
  \includegraphics[width= 0.9 \linewidth]{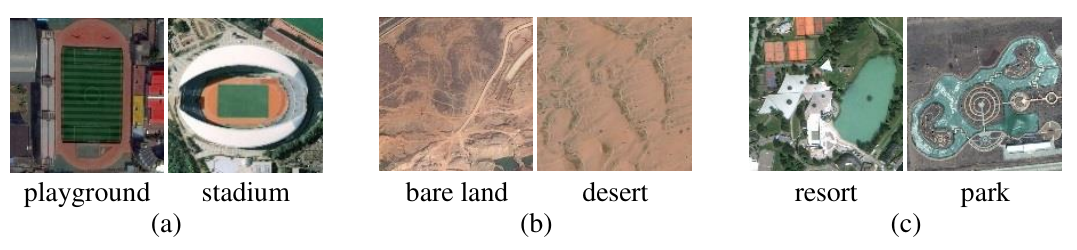}
  \caption{Small inter-class distance: (a). similar objects between different scenes; (b). similar textures between different scenes; (c). similar structural distributions between different scenes.}
  \label{inter}
  \end{figure}

  \item [-]\textbf{\emph{Relative large-scale dataset:}} For validating the classification algorithm, large-scale labeled data are often expected. However, the mannual annotation of aerial images requires expertise and is extremely time-consuming. AID has a total number of $10000$ images which is, to our knowledge, the largest annotated aerial image datasets. It can cover a much wider range of aerial images and better aproximate the real aerial image classification problem than existing dataset.
%In addition, the number of sample images in each class are also much larger than others, which can help to evaluate the classification performance more precisely.
In contrast with our AID, both the UC-Merced dataset~\cite{yang2010bag} and WHU-RS19 dataset~\cite{sheng2012high} contain $100$ images per class and only $20$ images in each class were usually used for testing the algorithms~\cite{yang2010bag,zhao2014land,zhu2014multi,chen2015pyramid,yang2011spatial,cheriyadat2014unsupervised,hu2015comparative,hu2015transferring,hu2015unsupervised,castelluccio2015land,penatti2015deep,yang2015learning,zhang2015saliency,zhang2015scene,zhu2015scene}. In such case, the classification accuracy will be seriously affected by even one image is predicted correctly or not and result in big standard deviation, especially when analyzing the classification results on each class.
 Therefore, our AID with relatively large-scale data can provide a better benchmark to evaluate image classification methods.

%  \item [-]\textbf{\emph{Variation of class size:}} The existing public datasets~\cite{yang2010bag,xia2010structural} are both uniformly distributed with the same number of images per class, which makes us difficult to judge whether the classification methods are biased to some specific classes for the Kappa coefficient is proportional to the overall accuracy (OA) under such a situation. However, Kappa coefficient is thought to be a more reliable measure than OA in remote sensing for it can take the classification accuracy of each class into account. Therefore, our new large scale dataset is designed with different number of images in each class (see Table.~\ref{tab:class}) to enable one to compare the Kappa coefficients of different methods so as to give a more comprehensive result.
\end{itemize}

%In all, the various properties make the new dataset to be more challenging, on which we can compare various algorithms more fairly and promote better algorithms.

\section{Baseline methods}
\label{sec:baseline}
\begin{figure*}[htb!]
  \centering
  \includegraphics[width=  0.8 \linewidth]{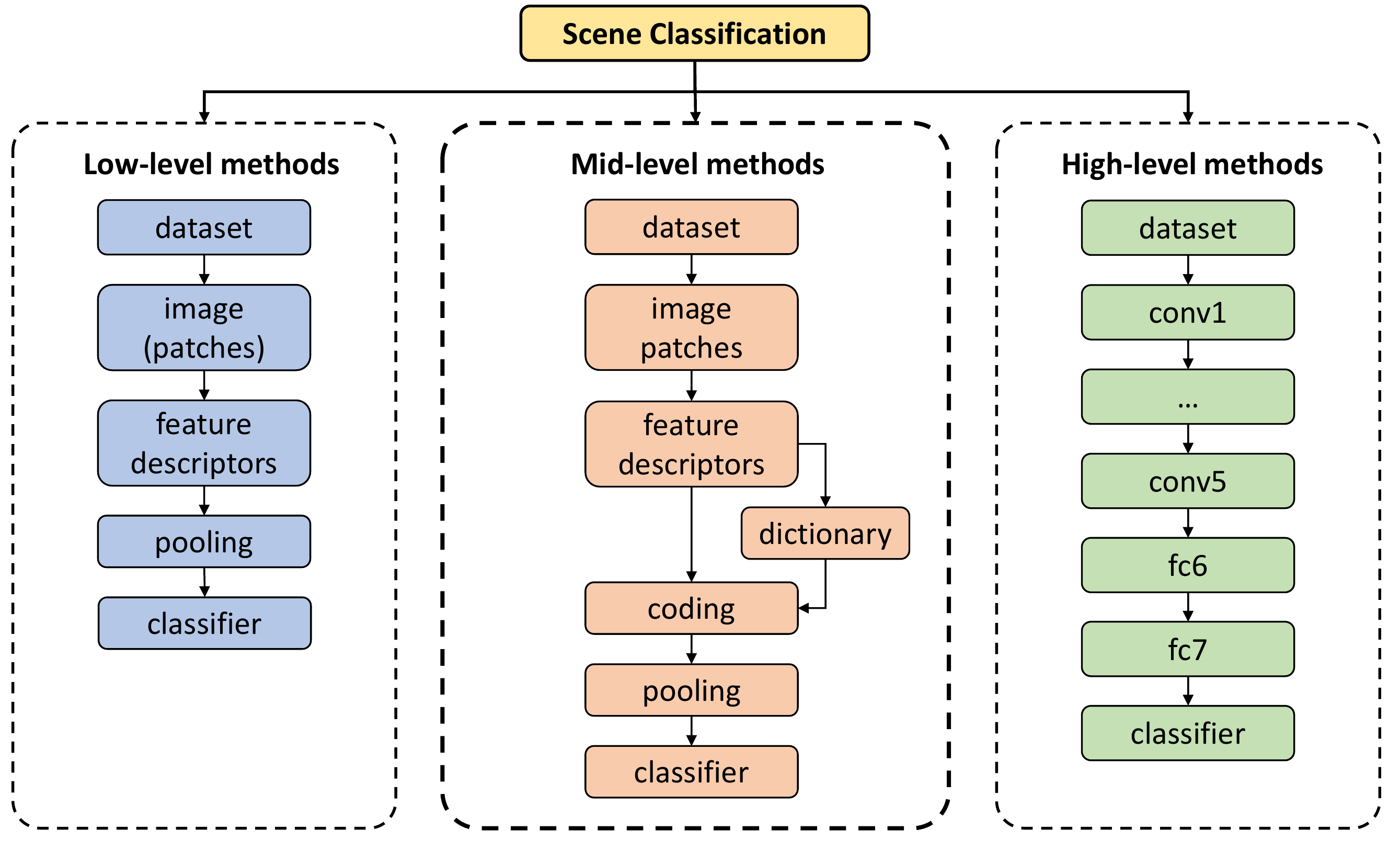}
  \caption{The general pipeline of three types of scene classification methods.}
  \label{method-classification}
\end{figure*}

In this section, we evaluate different aerial scene classification methods with low-, mid- and high-level scene descriptions reviewed previously\footnote{The codes of the baseline methods are downloadable at~\url{www.lmars.whu.edu.cn/xia/AID-project.html}.}. The general classification pipeline is demonstrated in Fig.~\ref{method-classification}.
%For each type, we choose some representative ones for evaluation, and the details of each methods will be illustrated in the following.

\subsection{Methods with low-level scene features}
%Early methods directly use the low-level visual features like spectral information, texture information or structure information etc. as image descriptors for aerial image classification.
Aerial image classification methods using low-level scene features often first partition an aerial image into small patches, then use low-level visual features, e.g. spectral information, texture information or structure information etc., to characterize patches and finally output the distribution of the patch features as the scene descriptor. In our tests, we choose four commonly used low-level methods in the experiment, {\em i.e.} SIFT~\cite{lowe2004sift}, LBP~\cite{ojala2002lbp}, Color Histogram (CH)~\cite{swain1991color} and GIST~\cite{oliva2001modeling}.

\begin{itemize}
  \item[-] SIFT~\cite{lowe2004sift}: It describes a patch by the histograms of gradients computed over a $4\times4$ spatial grid. The gradients are then quantized into eight bins so the final feature vector has a dimension of $128$ ($4\times4\times8$). %As it focuses on the histograms of gradient directions, thus can characterize the structure information of the patch well.
  \item[-] LBP~\cite{ojala2002lbp}: Some works adopt LBP to extract texture information from aerial images, see~\cite{dos2010evaluating,chen2011evaluation}. For a patch, it first compares the pixel to its $8$ neighbors: when the neighbor's value is less than the center pixel's, output ``1", otherwise, output ``0". This gives an $8$-bit decimal number to describe the center pixel.
 % binary number in each neighbor. After that, the 8 binary numbers are concatenated in turn to form a $8$-digit binary number and converted to a decimal number to describe the center pixel.
The LBP descriptor is obtained by computing the histogram of the decimal numbers over the patch and results in a feature vector with $256$ dimensions.
  \item[-] Color histogram~\cite{swain1991color}: Color histograms (CH) are used for extracting the spectral information of aerial scenes~\cite{dos2010evaluating,chen2011evaluation,yang2013geographic}. In our experiments, color histogram descriptors are computed separately in three channels of the RGB color space. Each channel is quantized into $32$ bins to form a total histogram feature length of $96$ by simply concatenation of the three channels.
  \item[-] GIST~\cite{oliva2001modeling}: Unlike aforementioned descriptors that focus on local information, GIST represents the dominant spatial structure of a scene by a set of perceptual dimensions (naturalness, openness, roughness, expansion, ruggedness) based on the spatial envelope model~\cite{oliva2001modeling} and thus widely used for describing scenes~\cite{risojevic2011gabor}. This descriptor is implemented by convolving the gray image with multi-scale (with the number of $S$) and multi-direction (with the number of $D$) Gabor filters on a $4\times4$ spatial grid. By concatenating the mean vector of each grid, we get the GIST descriptor of an image with $16 \times S \times D$ dimensions.
\end{itemize}

\subsection{Methods with mid-level scene features}
In contrast with low-level methods, mid-level aerial scene classification methods often build a scene representation by coding low-level local feature descriptors. In this paper, we evaluated $21$ commonly used mid-level features obtained by combining $3$ local feature descriptors (i.e., SIFT~\cite{lowe2004sift}, LBP~\cite{ojala2002lbp}and CH~\cite{swain1991color}) with $7$ mid-level feature coding approaches.
\begin{itemize}
  \item[-] {\em Bag of Visual Words} (BoVW)~\cite{sivic2003bow} model an image by leaving out the spatial information and representing it with the frequencies of local visual words~\cite{yang2010bag}. BoVW model and its variants are widely used in scene classification~\cite{chen2011evaluation,shao2013hierarchical,shao2013extreme,zhao2014land,
      zhao2014wavelet,chen2015pyramid,hu2015benchmark,hu2015comparative,sridharan2015bag}. The visual words are often produced by clustering local image descriptors to form a dictionary (with a given size $K$), e.g. using k-means algorithm.
      %Thus, by coding all the local patch descriptors of an image to the visual words in the dictionary using nearest neighbors, i.e., vector quantization (VQ), and counting the occurrence frequencies of the visual words, the global image feature is obtained with $K$ dimensions.

  \item[-] {\em Spatial Pyramid Matching} (SPM)~\cite{lazebnik2006spm} uses a sequence of increasingly coarser grids to build a spatial pyramid (with $L$ levels) coding of local image descriptors. By concatenating the weighted local image features in each subregion at different scales, one can get a $\frac{(4^{L}-1)\times{K}}{3}$ dimension global feature vector which is much longer than BoVW with the same size of dictionary~($K$).

  \item[-] {\em Locality-constrained Linear Coding} (LLC)~\cite{wang2010llc} is an effective coding scheme adapted from sparse coding methods~\cite{dai2011satellite,sheng2012high,zheng2013automatic}. It utilizes the locality constraints to code each local descriptor into its local-coordinate system by modifying the sparsity constraints~\cite{yang2009scspm,yu2009lcc}. The final feature can be generated by max pooling of the projected coordinates with the same size of dictionary.

  \item[-] {\em Probabilistic Latent Semantic Analysis} (pLSA)~\cite{bosch2006plsa} is a way to improve the BoVW model by topic models. A latent variable called topic is introduced and defined as the conditional probability distribution of visual words in the dictionary. It can serve as a connection between the visual words and  images. By describing an image with the distribution of topics (the number of topics is set to be $T$), one can solve the influence of synonym and polysemy meanwhile reduce the feature dimension to be $T$.

  \item[-] {\em Latent Dirichlet allocation} (LDA)~\cite{blei2003lda} is a generative topic model evolved from pLSA with the main difference that it adds a Dirichlet prior to describe the latent variable topic instead of the fixed Gaussian distribution, and is also widely used for scene classification~\cite{zhao2013hybrid,kusumaningrum2014integrated,zhong2015scene,zhu2015scene}. As a result, it can handel the problem of overfitting and also increase the robustness. The dimension of final feature vector is the same with the number of topics $T$.

  \item[-] {\em Improved Fisher kernel} (IFK)~\cite{Perronnin2010ifk}
  uses Gaussian Mixture Model (GMM) to encode local image features~\cite{perronnin2007fisher} and achieves good performance in scene classification~\cite{negrel2014evaluation,hu2015comparative,hu2015benchmark}.
  In essence, the feature of an image got by Fisher vector encoding method is a gradient vector of the log-likelihood. By computing and concatenating the partial derivatives of the mean and variance of the Gaussian functions, the final feature vector is obtained with the dimension of $2\times K \times F$ (where $F$ indicates the dimension of the local feature descriptors and $K$ denotes the size of the dictionary).
   %Compared with BoVW, it is not limited to the frequencies of each visual word but also encodes additional high-order information about the distribution of the descriptors.

  \item[-] {\em Vector of Locally Aggregated Descriptors} (VLAD)~\cite{jegou2012vlad} can be seen as a simplification of the IFK method which aggregates descriptors based on a locality criterion in feature space~\cite{negrel2014evaluation}. It uses the non-probabilistic k-means clustering to generate the dictionary by taking the place of GMM model in IFK. When coding each local patch descriptor to its nearest neighbor in the dictionary, the differences between them in each dimension are accumulated and resulting in an image feature vector with dimension of $K \times F$.
\end{itemize}

\subsection{Methods with high-level scene features}
In recent years, learned high-level deep features have been reported to achieve impressive results on aerial image classification~\cite{hu2015transferring,castelluccio2015land,penatti2015deep,luus2015multiview,zhang2015scene,zou2015deep,nogueira2016towards}. In this work, we also compare $3$ representative high-level deep-learned scene classification methods in our benchmark.

\begin{itemize}
  \item[-] CaffeNet: Caffe (Convolutional Architecture for Fast Feature Embedding)~\cite{jia2014caffe} is one of the most commonly used open-source frameworks for deep learning (deep convolutional neural networks in particular). The reference model - CaffeNet, which is almost a replication of ALexNet~\cite{krizhevsky2012imagenet} that is  proposed for the ILSVRC 2012 competition~\cite{ILSVRC15}. The main differences are: (1) there is no data argumentation during training; (2) the order of normalization and pooling are switched. Therefore, it has quite similar performances to the AlexNet, see~\cite{hu2015transferring,nogueira2016towards}. For this reason, we only test CaffeNet in our experiment. The architecture of CaffeNet comprises 5 convolutional layers, each followed by a pooling layer, and 3 fully connected layers at the end. In our work, we directly use the pre-trained model obtained using the ILSVRC 2012 dataset~\cite{ILSVRC15}, and extract the activations from the first fully-connected layer, which results in a vector of 4096 dimensions for an image.
%  \item[-] VGG-M: To conduct a rigorous empirical evaluation of CNN-based methods for image classification, ~\cite{chatfield2014vgg} explored three different deep architectures and comparing them on a common ground using Caffe toolkit. In our experiment, we choose the network which has a better speed/accuracy trade-off, named VGG-M, for evaluation. Similar to CaffeNet, it is also composed of 5 convolutional layers followed by 3 fully connected layers. To balance the speed/accuracy trade-off, it decreases the sizes of the stride and receptive field of the first convolutional layer, which is beneficial on the ILSVRC dataset. At the same time, the second convolutional layer uses larger stride to keep the computation time reasonable. As the same with CaffeNet, the activations from the first fully connected layer are extracted as feature representations.
  \item[-] VGG-VD-16: To investigate the effect of the convolutional network depth on its accuracy in the large-scale image recognition setting, \cite{Simonyan14vggvd} gives a thorough evaluation of networks by increasing depth using an architecture with very small ($3\times3$) convolution filters, which shows a significant improvement on the accuracies, and can be  generalised well to a wide range of tasks and datasets. In our work, we use one of its best-performing models, named VGG-VD-16, because of its simpler architecture and slightly better results. It is composed of 13 convolutional layers and followed by 3 fully connected layers, thus results in 16 layers. Similarly, we extract  the activations from the first fully connected layer as the feature vectors of the images.
  \item[-] GoogLeNet: This model~\cite{szegedy2014going} won the ILSVRC-2014 competition~\cite{ILSVRC15}. Its main novelty lies in the design of the "Inception modules", which is based on the idea of "network in network"~\cite{lin2013network}. By using the Inception modules, GoogLeNet has two main advantages: (1) the utilization of filters of different sizes at the same layer can maintain multi-scale spatial information; (2) the reduction of the number of parameters of the network makes it less prone to overfitting and allows it to be deeper and wider. Specifically, GoogLeNet is a 22-layer architecture with more than 50 convolutional layers distributed inside the inception modules. Different from the above CNN models,
      GoogLeNet has only one fully connected layer at last, therefore, we extract the features of the fully connected layer for testing.
\end{itemize}

\section{Experimental studies}
\label{sec:expertment}
We evaluate all the three kinds of scene classification methods mentioned before: methods with low-level, mid-level and high-level scene features. For each type, we choose some representative ones as baseline for evaluation: SIFT~\cite{lowe2004sift}, LBP~\cite{ojala2002lbp}~\cite{ojala2002lbp}, Color Histogram (CH)~\cite{swain1991color} and GIST~\cite{oliva2001modeling} for low-level methods, BoVW~\cite{sivic2003bow}, Spatial Pyramid Matching (SPM)~\cite{lazebnik2006spm}, Locality-constrained Linear Coding (LLC)~\cite{wang2010llc}, Probabilistic Latent Semantic Analysis (pLSA)~\cite{bosch2006plsa}, Latent Dirichlet allocation (LDA)~\cite{blei2003lda}, Improved Fisher kernel (IFK)~\cite{Perronnin2010ifk} and Vector of Locally Aggregated Descriptors (VLAD)~\cite{jegou2012vlad} combined with three local feature descriptors (i.e., SIFT~\cite{lowe2004sift}, LBP~\cite{ojala2002lbp}, CH~\cite{swain1991color}) for mid-level methods, and three representative high-level deep-learning methods (i.e., CaffeNet~\cite{jia2014caffe}, VGG-VD-16\cite{Simonyan14vggvd} and GoogLeNet~\cite{szegedy2014going}) are adopted.

\subsection{Parameter Settings}
In our experiment, we firstly test four kinds of low-level methods for classification: SIFT, LBP, CH and Gist. For the local patch descriptor SIFT, we use a fixed size grid ($16\times16$ pixels) with the spacing step to be 8 pixels to extract all the descriptors in the gray image plain and adopt the average pooling method for each dimension of the descriptor so as to get the final image feature with 128 dimensions (8 orientations with $4\times4$ subregions). As for the remaining three descriptors, we use them as global descriptors that can extract the feature vectors on the whole image very efficiently. For LBP, we use the common used 8 neighbors to get the binary values and convert the 8-bit binary values into a decimal value for each pixel in the gray image. By computing the frequencies of the 256 patterns, we get the low-level LBP features of an image. For CH, we directly use the RGB color space and quantize each channel into 32 bins. Thus, the feature of an image is obtained by concatenation of the statistical histograms in each channel and result in 96 dimensions. For Gist descriptor, we set the same parameters as in its original work~\cite{oliva2001modeling}: the number of scales is set to be 4, the orientations are quantized into 8 bins and a $4\times4$ spatial grid is utilized for pooling, thus, it results in 512 dimensions ($4\times8\times4\times4$).

For mid-level methods, we test afore-mentioned seven different feature coding methods: BoVW, SPM, LLC, pLSA, LDA, IFK and VLAD. Three local patch descriptors - SIFT, LBP and CH have been utilized for extracting the local structure, texture and spectral features respectively. In the patch sampling procedure, we use the grid sampling as our previous work~\cite{hu2015comparative} has proven that grid sampling has better performance for scene classification of remote sensing imagery. Therefore, we set the patch size to be $16\times16$ pixels and the grid spacing to be 8 pixels for all the local descriptors to balance the speed/accuracy trade-off. By combining the three local feature descriptors and seven global feature coding methods, we can get 21 different mid-level features in all. As for the size of the dictionary, we set it from 16 to 8192 for the 7 coding methods when using SIFT as local feature descriptors, and select the optimal one when using LBP and CH for describing local patches. For some special parameters defined in each coding methods, we empirically set the spatial pyramid level to be 2 in SPM, and both the numbers of topics in pLSA and LDA are set to be a half of the dictionary size.

For high-level methods, we just use the CNN models pre-trained on the ILSVRC 2012 dataset~\cite{ILSVRC15} and extract the features from the first fully connected layer in each CNN model as the global features. CaffeNet and VDD-VD-16 result in a vector of 4096 dimensions while GoogLeNet a 1024-dimensional feature vector owing to the fact that GoogLeNet has only one fully connected layer, and all the features are $L2$ normalized for better performance.

After getting the global features using various methods, we use the liblinear~\cite{Fan2010LIBLINEAR} for supervised classification because it can quickly train a linear classifier on large scale datasets. More specifically, we spilt the images in the dataset into training set and testing set. The features of the training set are used to training a linear classification model by liblinear, and the features of the testing set are used for estimating the performance of the trained model.

\subsection{Evaluation protocols}

To compare the classification quantitatively, we compute the common used measures: overall accuracy (OA) and Confusion matrix. OA is defined as the number of correctly predicted images divided by the total number of predicted images. It is a direct measure to reveal the classification performance on the whole dataset. Confusion matrix is a specific table layout that allows direct visualization of the performance on each class. Each column of the matrix represents the instances in a predicted class, and each row represents the instances in an actual class, thus, each item ${x}_{ij}$ in the matrix computes the proportion of images that predicted to be the $i$-th type meanwhile trully belong to the $j$-th type.

To compute OA, we adopt two different settings for each tested dataset in the supervised classification process. For the RSSCN7 dadaset and our AID dataset, we fix the ratio of the number of training set to be 20\% and 50\% respectively and the left for testing, while for UC-Merced dataset, the ratios are set to be 50\% and 80\% respectively. For the WHU-RS19 dataset, the ratios are fixed at 40\% and 60\% respectively. To compute the overall accuracy, we randomly split the datasets into training sets and testing sets for evaluation, and repeat it ten times to reduce the influence of the randomness and obtain reliable results. The OA is computed for each run, and the final result is reported as the mean and standard deviation of OA from the individual run.

To compute the confusion matrix, we fix the training set by choosing the same images for fair comparison on each datasets and fix the ratio of the number of training set of the UC-Merced dataset, the WHU-RS19 dataset,the RSSCN7 dadaset and our AID dataset to be 50\%, 40\%, 20\% and 20\% respectively.

\subsection{Experimental results}

In this section, we evaluate different methods on the common used UC-Merced dataset, WHU-RS19 dataset, RSSCN7 dataset as well as our AID dataset, and give the corresponding results and analysis, which are divided into four phases: results of low-level methods, results of mid-level methods, results of high-level methods and confusion matrix.

\subsubsection{Results with low-level methods}

\begin{table*}[htb!]
\renewcommand\arraystretch{1.5}
\caption{Overall accuracy (OA) of different low-level methods on the UC-Merced dataset, the WHU-RS19 dataset, the RSSCN7 dataset and our AID dataset.}
\label{low-level-oa}
\centering
\tiny
\begin{tabular}{c|c|c|c|c|c|c|c|c}
\Xhline{1.5pt}
 Methods & UC-Merced (.5) &UC-Merced (.8) & WHU-RS19 (.4) & WHU-RS19 (.6) & RSSCN7 (.2)  &  RSSCN7 (.5)   & AID (.2) & AID (.5) \\ \hline
 SIFT    & 28.92$\pm$0.95   &32.10$\pm$1.95   & 25.37$\pm$1.32  & 27.21$\pm$1.77  & 28.45$\pm$1.03 &  32.76$\pm$1.25  & 13.50$\pm$0.67     & 16.76$\pm$0.65 \\
 LBP     & 34.57$\pm$1.38   &36.29$\pm$1.90   & 40.11$\pm$1.46  & 44.08$\pm$2.02  & 57.55$\pm$1.18 &  60.38$\pm$1.03  & 26.26$\pm$0.52     & 29.99$\pm$0.49 \\
 CH      & 42.09$\pm$1.14   &46.21$\pm$1.05   & 48.79$\pm$2.37  & 51.87$\pm$3.40  & 57.20$\pm$1.23 &  60.54$\pm$1.01  & 34.29$\pm$0.40     & 37.28$\pm$0.46 \\
 GIST    & 44.36$\pm$1.58   &46.90$\pm$1.76   & 45.65$\pm$1.06  & 48.82$\pm$3.12  & 49.20$\pm$0.63 &  52.59$\pm$0.71  & 30.61$\pm$0.63     & 35.07$\pm$0.41 \\
\Xhline{1.5pt}
\end{tabular}
\end{table*}

Table.~\ref{low-level-oa} illustrates the means and standard variances of OA using the four kinds of low-level methods (e.g. SIFT, LBP, CH, GIST) with randomly choosing the fixed percent of images to construct the training set by repeating 10 times on the UC-Merced dataset, the WHU-RS19 dataset, the RSSCN7 dataset and our AID dataset. Although different features give different performance on different datasets, we can observe the consistent phenomenon on all datasets that SIFT descriptor performs far less than others with about 20\% lower OA than the highest ones, which indicates that SIFT descriptor is not suitable to be as low-level feature for directly classification. For the other three low-level features, GIST performs the best on the UC-Merced dataset, and CH gives the best performances on both WHU-RS19 dataset and our AID dataset, while LBP and CH give comparable results on the RSSCN7 dataset. The different performances can be explained by the characteristics of the datasets, for example, both the UC-Merced dataset and our AID dataset contain various artificial scene types, which are mainly made up of various buildings, therefore, GIST, which can extract the dominant spatial structure of a scene, performs well on these datasets. For the RSSCN7 datasets, which contains much natural scene types, thus, the texture feature descriptor LBP works the best. In addition, CH gives the most robust performances on all the datasets, because most scene types are color consistent, e.g., grass ia mostly green and desert is dark yellow.

%Note that the standard deviations of OA on our new dataset are much lower than the other two datasets, which is caused by the number of testing samples. There are only 20 images per class for testing of the UC-Merced dataset and WHU-RS19 dataset, thus, OA will have a greater variation range if the numbers of right predictions in each run are inconsistent with only a few numbers. But the number of testing images in our new dataset is more than 10 times larger than the other two, thus, it will result in much smaller standard variances, which can help to evaluate the performances more precisely.

\subsubsection{Results with mid-level methods}

\begin{figure*}[htb!]
\centering
%begin{tabular}{cc}
\subfigure[UC-Merced]{\includegraphics[width= 0.42\linewidth]{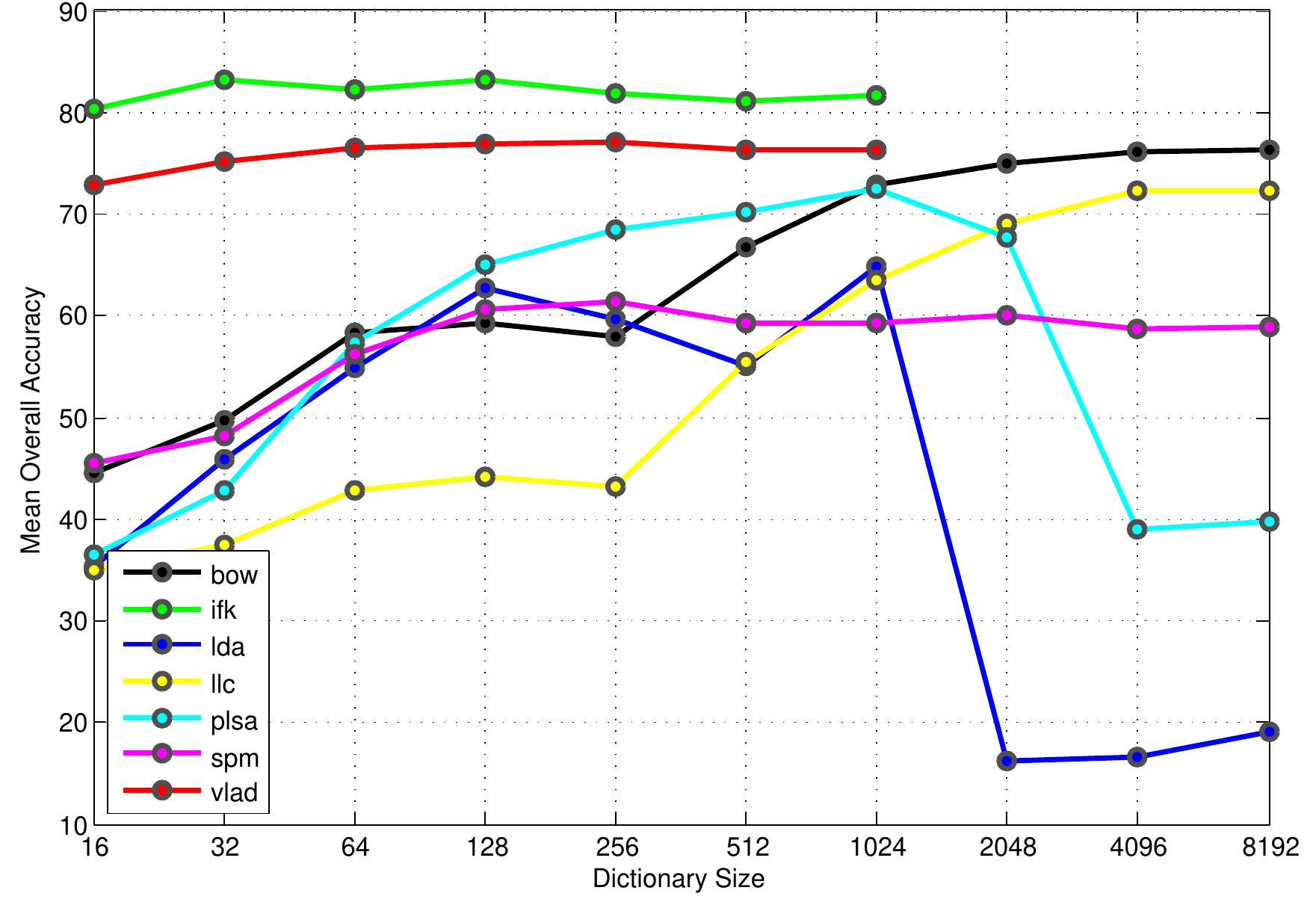}}
\subfigure[WHU-RS19]{\includegraphics[width= 0.42\linewidth]{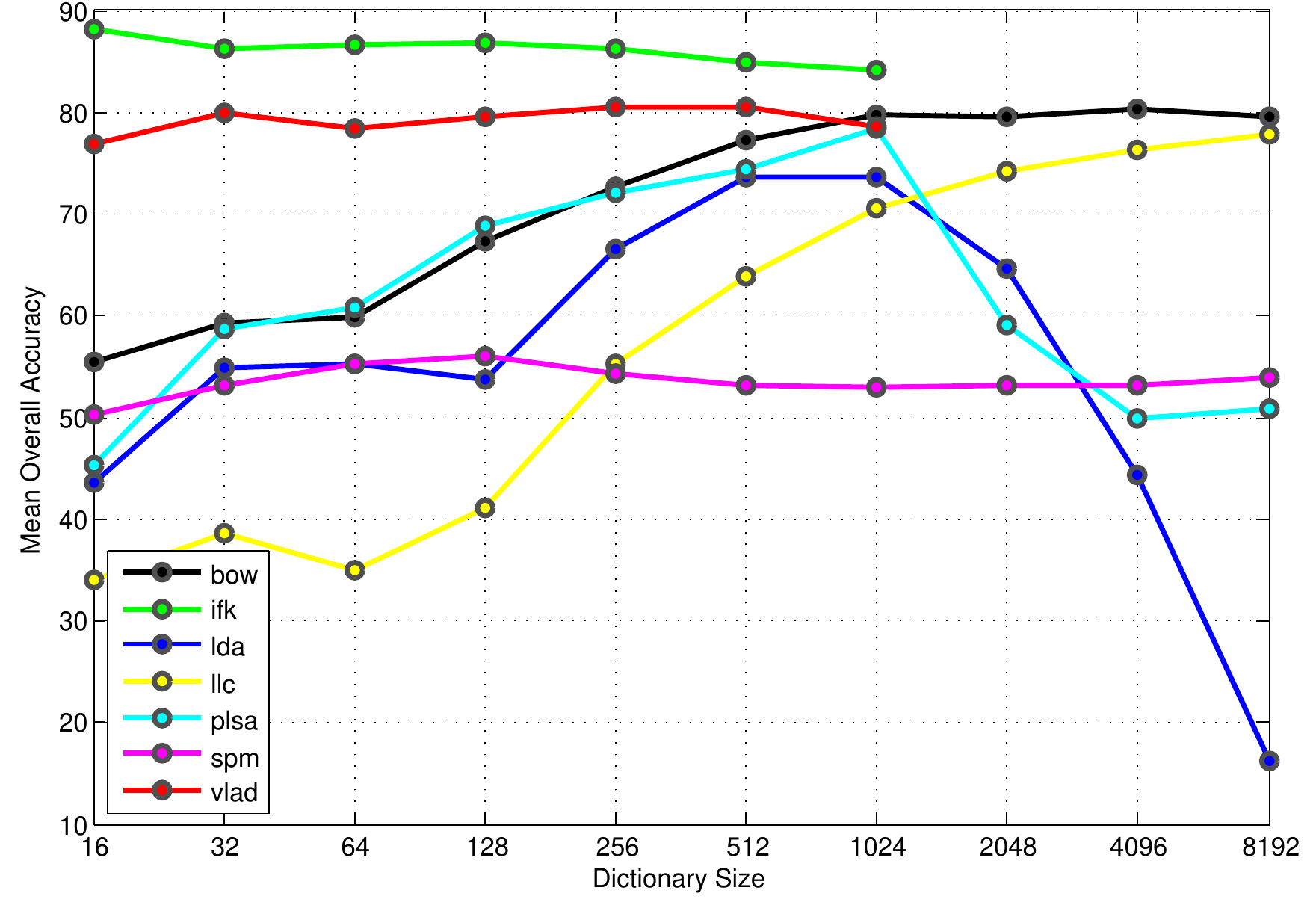}}\\
\subfigure[RSSCN7]{\includegraphics[width= 0.42\linewidth]{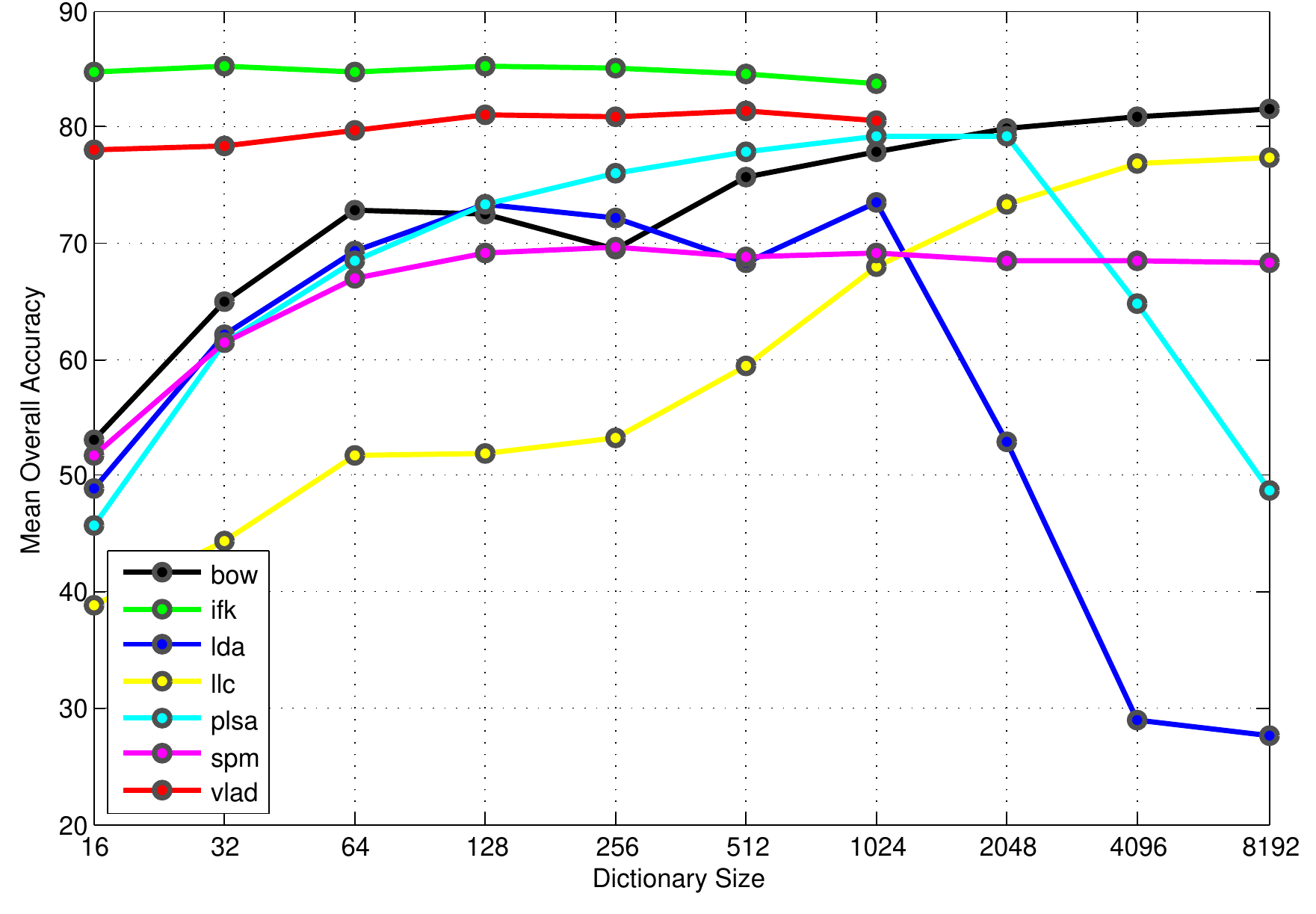}}
\subfigure[AID]{\includegraphics[width= 0.42\linewidth]{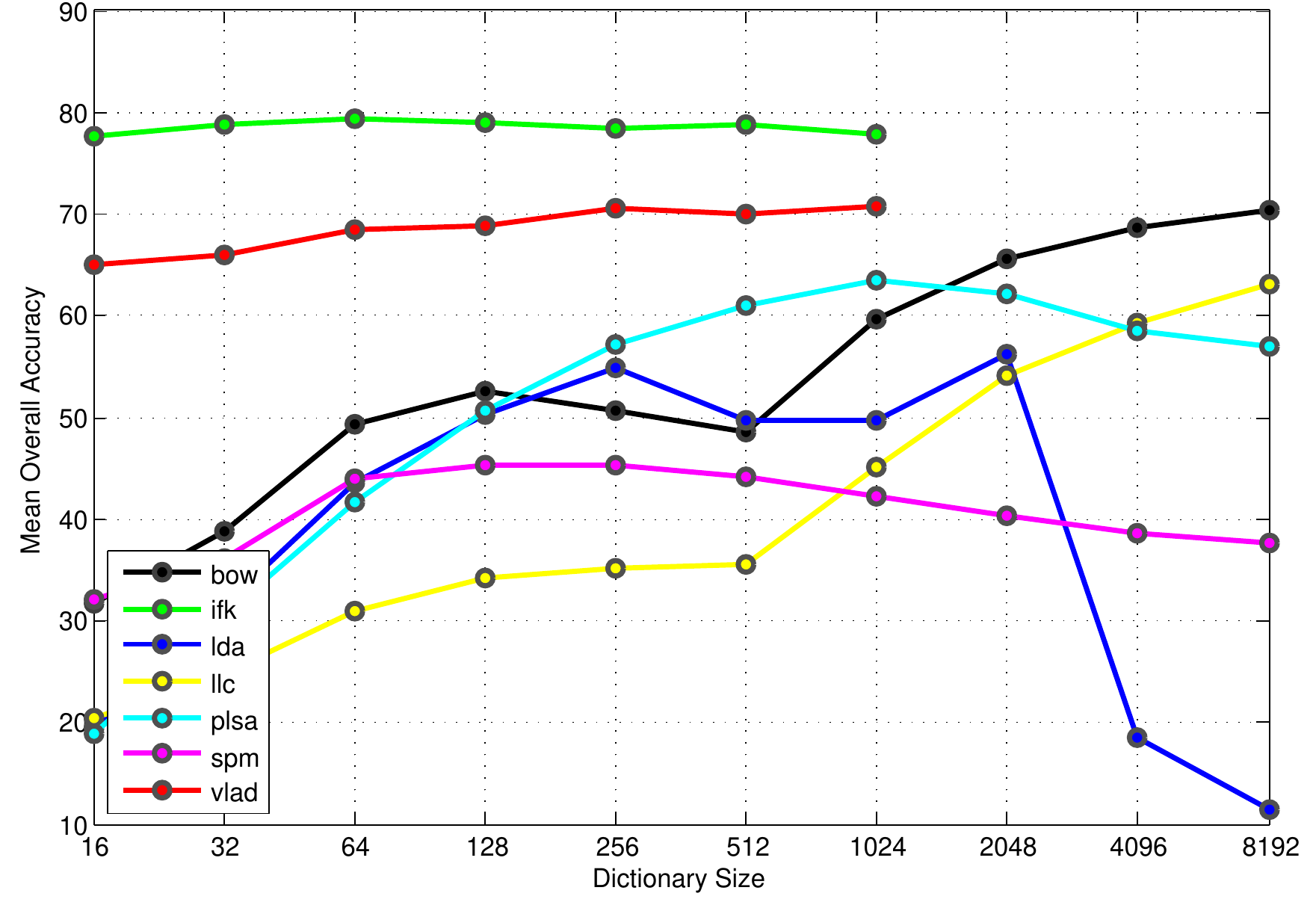}}
%end{tabular}
\caption{Overall accuracy (OA) of different mid-level coding methods using different dictionary size on the UC-Merced dataset, the WHU-RS19 dataset and our AID dataset.}
\label{dic}
\end{figure*}

For the seven kinds of feature coding methods, the size of the dictionary has a great influence on the classification results, therefore, we need to firstly find the optimal dictionary size for each coding method. To do so, we fix the local patch descriptor using SIFT, and gradually double increase the dictionary size from 16 to 8192 for each coding methods on the three datasets. The corresponding OA is shown in Fig.~\ref{dic}. For BoVW, the larger the dictionary, the better the performance. However, the performance increases quite slowly when the dictionary size becomes larger. Therefore, we fix the dictionary size of BoVW to be 4096 in the following experiment. For IFK and VLAD, the dictionary size has quite little influence on the performance for all the datasets. But the larger dictionary will result in much higher dimensional features and thus more time-consuming for training classification model, therefore, the corresponding dictionary sizes are fixed at 32 and 64 respectively. For LDA, pLSA and SPM, there is a drop when the size of the dictionary achieves to some degree. Thus, we choose 1024, 1024 and 128 to be the corresponding dictionary sizes which are suitable for all the datasets. For LLC, the performance is increasing all the way with the size of the dictionary, therefore, we fix it at 8192 in the following experiment. Note that the parameters we choose are not always the optimal ones, which is to make a trade-off between accuracy and speed.

\begin{table*}[htb!]
\renewcommand\arraystretch{1.5}
\caption{Overall accuracy (OA) of different mid-level methods on the UC-Merced dataset, the WHU-RS19 dataset, the RSSCN7 dataset and our AID dataset.}
\label{mid-level-oa}
\centering
\tiny
\begin{tabular}{c|c|c|c|c|c|c|c|c}
\Xhline{1.5pt}
 Methods & UC-Merced (.5) &UC-Merced (.8) & WHU-RS19 (.4) & WHU-RS19 (.6) & RSSCN7 (.2)  &  RSSCN7 (.5)   & AID (.2) & AID (.5) \\ \hline
  BoVW (SIFT) & 72.40$\pm$1.30   & 75.52$\pm$2.13  & 77.21$\pm$1.92  & 82.58$\pm$1.72  & 76.91$\pm$0.59 & 81.28$\pm$1.19  & 62.49$\pm$0.53     & 68.37$\pm$0.40 \\
  IFK (SIFT)  & 78.74$\pm$1.65   & 83.02$\pm$2.19  & 83.35$\pm$1.19  & 87.42$\pm$1.59  & 81.08$\pm$1.21 & 85.09$\pm$0.93  & 71.92$\pm$0.41     & 78.99$\pm$0.48 \\
  LDA (SIFT)  & 59.24$\pm$1.66   & 61.29$\pm$1.97  & 69.91$\pm$2.23  & 72.18$\pm$1.58  & 71.07$\pm$0.70 & 73.86$\pm$0.77  & 51.73$\pm$0.73     & 50.81$\pm$0.54 \\
  LLC (SIFT)  & 70.12$\pm$1.09   & 72.55$\pm$1.83  & 73.28$\pm$1.37  & 78.63$\pm$2.04  & 73.29$\pm$0.63 & 77.11$\pm$1.29  & 58.06$\pm$0.50     & 63.24$\pm$0.44 \\
  pLSA (SIFT) & 67.55$\pm$1.11   & 71.38$\pm$1.77  & 73.25$\pm$1.80  & 77.50$\pm$1.20  & 75.25$\pm$1.20 & 79.37$\pm$0.97  & 56.24$\pm$0.58     & 63.07$\pm$0.48 \\
  SPM (SIFT)  & 56.50$\pm$1.00   & 60.02$\pm$1.06  & 51.82$\pm$1.63  & 55.82$\pm$1.95  & 64.97$\pm$0.79 & 68.45$\pm$1.01  & 38.43$\pm$0.51     & 45.52$\pm$0.61 \\
  VLAD (SIFT) & 71.94$\pm$1.36   & 75.98$\pm$1.60  & 73.96$\pm$2.22  & 79.16$\pm$1.71  & 74.30$\pm$0.74 & 79.34$\pm$0.71  & 61.04$\pm$0.69     & 68.96$\pm$0.58 \\
  BoVW (LBP)  & 73.48$\pm$1.39   & 78.12$\pm$1.38  & 71.11$\pm$2.72  & 75.89$\pm$2.40  & 76.74$\pm$0.82 & 81.40$\pm$1.09  & 56.98$\pm$0.55     & 64.31$\pm$0.41 \\
  IFK (LBP)   & 73.11$\pm$1.08   & 78.02$\pm$1.60  & 71.02$\pm$2.66  & 75.61$\pm$1.86  & 75.18$\pm$1.18 & 80.31$\pm$1.46  & 60.11$\pm$0.56     & 69.22$\pm$0.72 \\
  LDA (LBP)   & 61.87$\pm$1.92   & 63.40$\pm$2.05  & 62.93$\pm$2.27  & 67.37$\pm$2.35  & 70.47$\pm$0.87 & 73.63$\pm$0.91  & 43.22$\pm$0.53     & 41.51$\pm$0.76 \\
  LLC (LBP)   & 67.19$\pm$1.40   & 72.95$\pm$1.46  & 72.89$\pm$1.98  & 76.00$\pm$0.99  & 73.28$\pm$0.56 & 77.46$\pm$0.86  & 56.11$\pm$0.61     & 61.53$\pm$0.55 \\
  pLSA (LBP)  & 68.84$\pm$1.18   & 74.07$\pm$1.71  & 66.07$\pm$2.20  & 71.08$\pm$2.11  & 74.94$\pm$0.52 & 78.97$\pm$1.19  & 49.71$\pm$0.55     & 57.31$\pm$0.58 \\
  SPM (LBP)   & 55.26$\pm$1.34   & 60.52$\pm$1.46  & 52.72$\pm$0.98  & 56.18$\pm$2.43  & 68.05$\pm$1.30 & 71.26$\pm$0.96  & 38.33$\pm$0.82     & 44.16$\pm$0.43 \\
  VLAD (LBP)  & 69.02$\pm$0.94   & 74.83$\pm$2.02  & 65.02$\pm$2.80  & 70.29$\pm$1.91  & 72.63$\pm$0.93 & 77.41$\pm$1.42  & 53.15$\pm$0.61     & 61.19$\pm$0.49 \\
  BoVW (CH)   & 69.80$\pm$1.11   & 76.33$\pm$2.32  & 63.26$\pm$1.52  & 67.29$\pm$1.55  & 75.07$\pm$1.18 & 81.74$\pm$0.60  & 49.16$\pm$0.24     & 56.84$\pm$0.45 \\
  IFK (CH)    & 73.87$\pm$1.09   & 79.14$\pm$1.91  & 70.04$\pm$1.47  & 74.89$\pm$2.25  & 76.86$\pm$0.78 & 83.32$\pm$0.72  & 59.60$\pm$0.66     & 67.49$\pm$0.81 \\
  LDA (CH)    & 60.15$\pm$1.23   & 64.12$\pm$1.75  & 55.23$\pm$1.57  & 58.92$\pm$2.56  & 68.11$\pm$1.85 & 71.29$\pm$1.30  & 38.16$\pm$0.60     & 41.70$\pm$0.80 \\
  LLC (CH)    & 68.62$\pm$1.68   & 73.00$\pm$1.41  & 64.46$\pm$1.68  & 68.82$\pm$1.94  & 74.12$\pm$0.80 & 79.94$\pm$0.92  & 53.47$\pm$0.43     & 58.23$\pm$0.49 \\
  pLSA (CH)   & 67.66$\pm$0.92   & 72.88$\pm$2.14  & 59.88$\pm$1.87  & 63.34$\pm$1.93  & 73.69$\pm$1.60 & 78.79$\pm$1.02  & 48.35$\pm$0.31     & 55.70$\pm$0.52 \\
  SPM (CH)    & 53.56$\pm$0.94   & 57.17$\pm$1.72  & 54.05$\pm$1.38  & 56.39$\pm$1.67  & 64.86$\pm$1.26 & 68.24$\pm$0.71  & 39.60$\pm$0.56     & 44.01$\pm$0.41 \\
  VLAD (CH)   & 67.69$\pm$1.58   & 72.48$\pm$2.24  & 59.53$\pm$1.85  & 63.97$\pm$2.32  & 72.59$\pm$1.02 & 79.21$\pm$0.87  & 47.94$\pm$0.39     & 57.34$\pm$0.73 \\

\Xhline{1.5pt}
\end{tabular}
\end{table*}

After finding the proper dictionary size for each coding method, we set the corresponding values for other local patch descriptors and evaluate the 21 kinds of mid-level features obtained by combining seven kinds of global feature coding methods (e.g., BoVW, SPM, LLC, pLSA, LDA, IFK, VLAD) with three kinds of local feature descriptors (e.g., SIFT, LBP, CH). Table.~\ref{mid-level-oa} shows the means and standard variances of OA on each dataset. Surprisingly, when comparing the results using different local feature descriptors, SIFT can give consistent the best performances, while CH performs the worst, while in the low-level features, SIFT is the worst among the low-level methods and CH is the most robust. This indicates that SIFT is more suitable to be encoded in the mid-level methods to generate more robust feature representation. By comparing the results using different global feature coding methods, BoVW and IFK account for the highest two OA in general, pLSA, LLC and VLAD are in the middle, while the left two methods have relatively worse performances. When comparing all the 21 mid-level methods, the features obtained by IFK with SIFT descriptor perform the best on all the datasets, which benefits from the combination of the great robustness and invariance of SIFT when describing local patches and the generative and discriminative nature of IFK.

\subsubsection{Results with high-level methods}

\begin{table*}[htb!]
\renewcommand\arraystretch{1.5}
\caption{Overall accuracy (OA) of high-level methods on the UC-Merced dataset, the WHU-RS19 dataset, the RSSCN7 dataset and our AID dataset.}
\label{high-level-oa}
\centering
\tiny
\begin{tabular}{c|c|c|c|c|c|c|c|c}
\Xhline{1.5pt}
 Methods & UC-Merced (.5) &UC-Merced (.8) & WHU-RS19 (.4) & WHU-RS19 (.6) & RSSCN7 (.2)  &  RSSCN7 (.5)   & AID (.2) & AID (.5) \\ \hline
  CaffeNet    & 93.98$\pm$0.67   & 95.02$\pm$0.81  & 95.11$\pm$1.20  & 96.24$\pm$0.56  & 85.57$\pm$0.95 & 88.25$\pm$0.62  & 86.86$\pm$0.47     & 89.53$\pm$0.31 \\
  VGG-VD-16   & 94.14$\pm$0.69   & 95.21$\pm$1.20  & 95.44$\pm$0.60  & 96.05$\pm$0.91  & 83.98$\pm$0.87 & 87.18$\pm$0.94  & 86.59$\pm$0.29     & 89.64$\pm$0.36 \\
  GoogLeNet   & 92.70$\pm$0.60   & 94.31$\pm$0.89  & 93.12$\pm$0.82  & 94.71$\pm$1.33  & 82.55$\pm$1.11 & 85.84$\pm$0.92  & 83.44$\pm$0.40     & 86.39$\pm$0.55 \\
\Xhline{1.5pt}
\end{tabular}
\end{table*}

Table.~\ref{high-level-oa} illustrates the means and standard variances of OA using the high-level methods (i.e., the features extracted from the first fully connected layer using the pre-trained CNN models) on the four datasets. From the classification results, we can see that CaffeNet and VGG-VD-16 give similar performances on all the datasets, while GoogLeNet performs slightly worse. Note that CaffeNet has only 8 layers that is much shallower than the VGG-VD-16 and GoogLeNet which has 16 and 22 layers respectively. Superficially, this phenomenon may result in the conclusion that shallower network works better, which is inconsistent with image classification of natural images. However, note the fact that the networks are all trained by the natural images, we just use them as feature extractors in our experiment. Therefore, the deeper the network, the more likely the learned features oriented to the natural image processing task, which may result in worse performance for classifying aerial scenes.

Compared with the above low-level and mid-level methods, high-level methods show far better performance on both datasets, which indicates that the high-level methods have the ability to learn highly discriminative features. Moreover, note that all the networks we use are pre-trained models on the ILSVRC 2012 dataset~\cite{ILSVRC15}, i.e., all the parameters are trained by the natural images, which shows its great generalization ability compared with other methods.

In addition, in all the above methods, the standard deviations of OA on our new dataset are much lower than the others, which is mainly caused by the number of testing samples. There are only dozens of images per class for testing of the UC-Merced dataset and WHU-RS19 dataset, thus, OA will have a greater variation range if the numbers of right predictions in each run are inconsistent with only a few numbers. But the number of testing images in our new dataset is more than 10 times larger than the above two, thus, it will result in much smaller standard variances, which can help to evaluate the performances more precisely.

\subsubsection{Confusion matrix}

\begin{figure*}[htb!]
\centering
\begin{tabular}{ccc}
\begin{minipage}[t]{0.33\linewidth}
 \centerline{\includegraphics[width=  \linewidth]{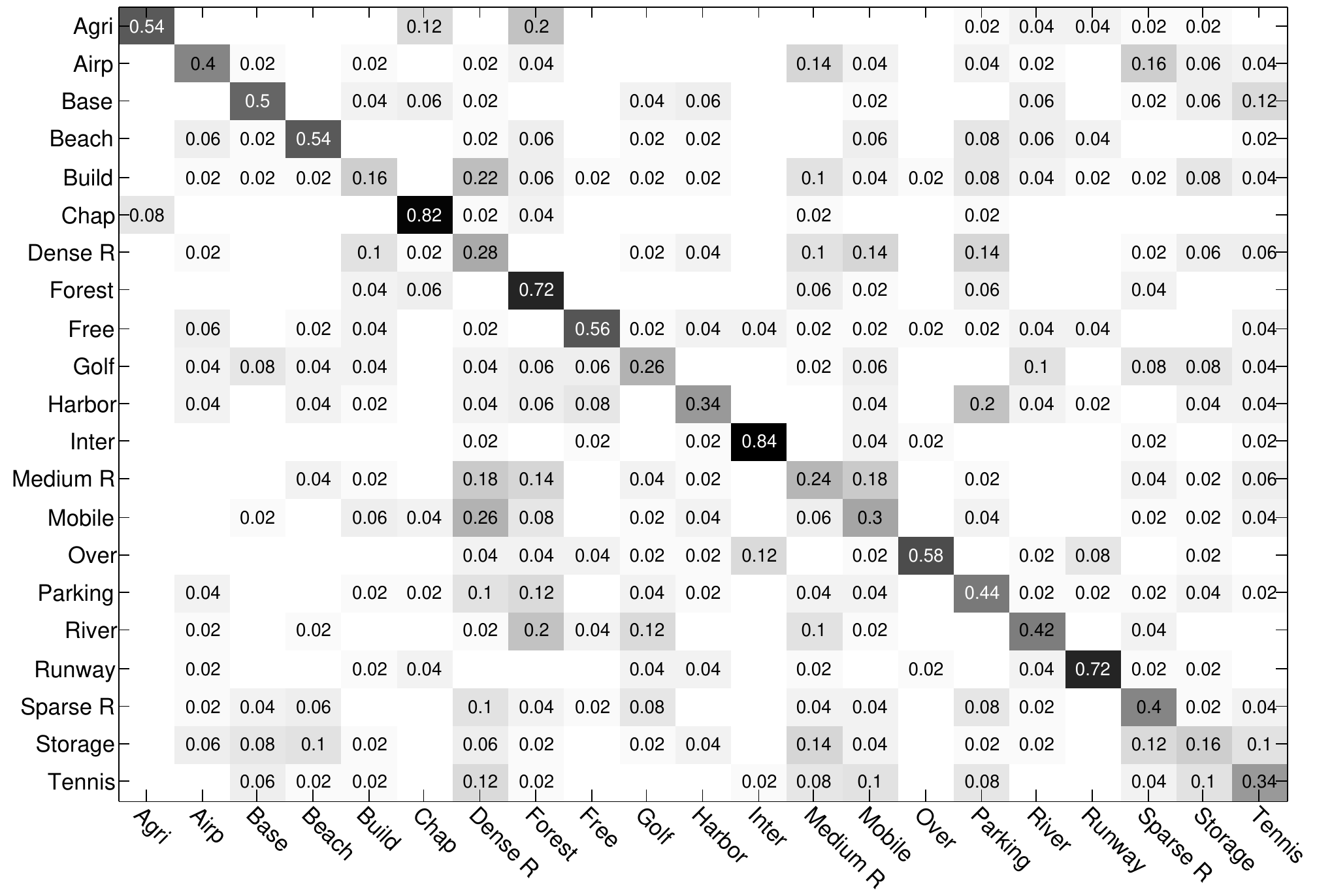}}
 \centerline{(a) low-level (GIST)}
\end{minipage}
\hfill
\begin{minipage}[t]{0.33\linewidth}
 \centerline{\includegraphics[width= \linewidth]{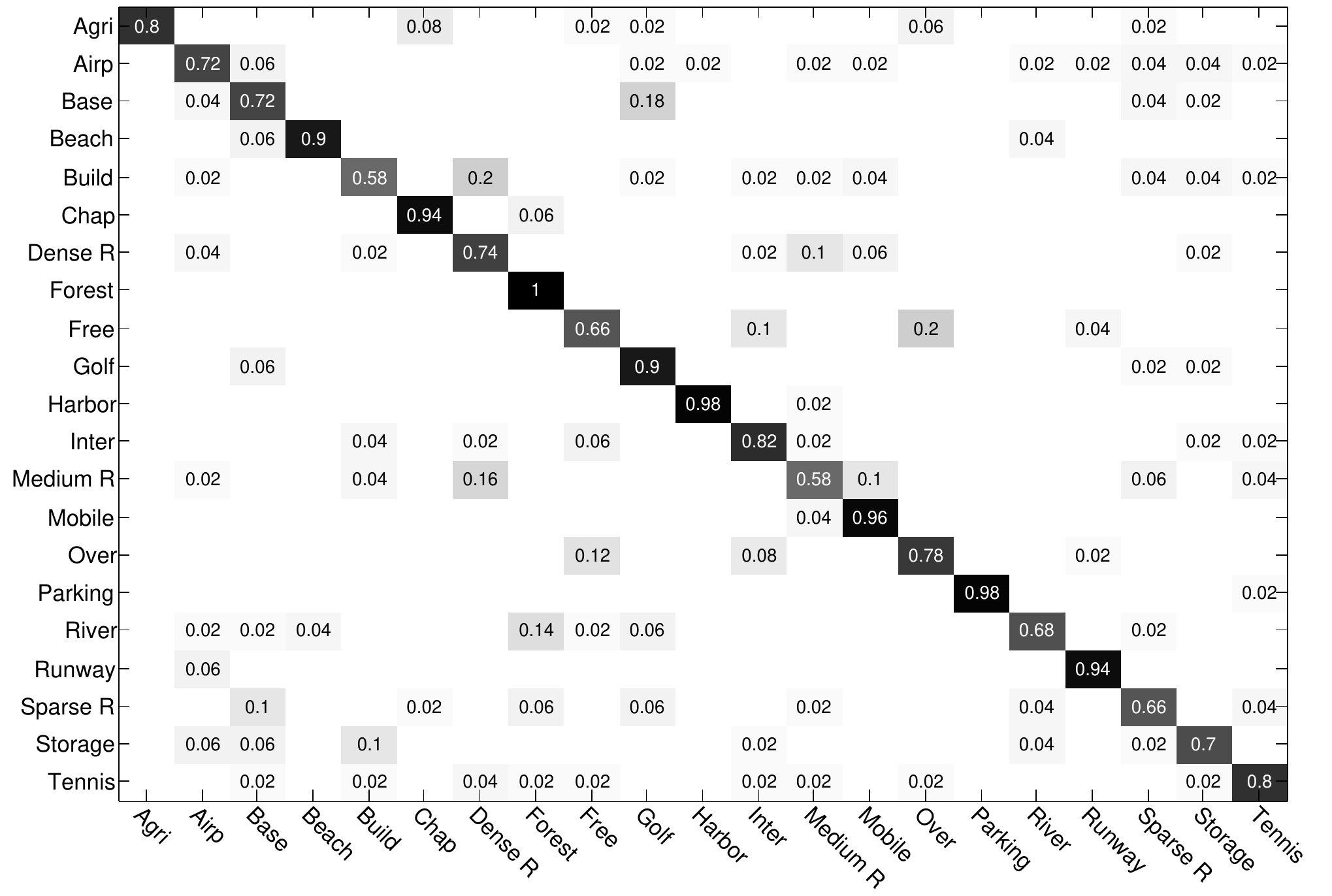}}
 \centerline{(b) mid-level IFK (SIFT)}
\end{minipage}
\hfill
\begin{minipage}[t]{0.33\linewidth}
 \centerline{\includegraphics[width=  \linewidth]{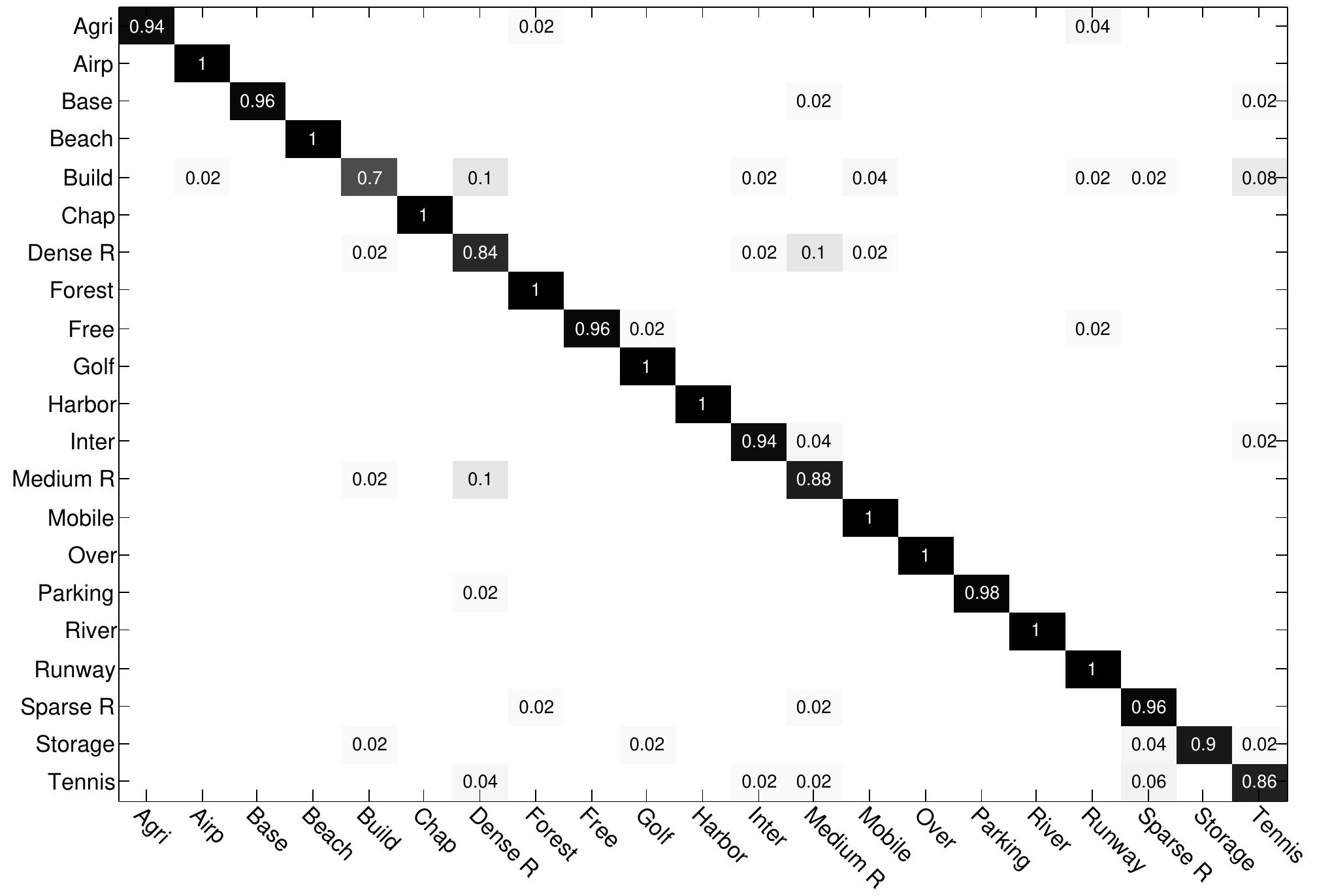}}
 \centerline{(b) high-level (VGG-VD-16)}
\end{minipage}
\end{tabular}
\caption{Confusion matrix obtained by
  low-level (GIST), mid-level IFK (SIFT) and high-level (VGG-VD-16) on the UC-Merced dataset.}
\label{ucm-cm}
\end{figure*}

\begin{figure*}[htb!]
\centering
\begin{tabular}{ccc}
\begin{minipage}[t]{0.33\linewidth}
 \centerline{\includegraphics[width=  \linewidth]{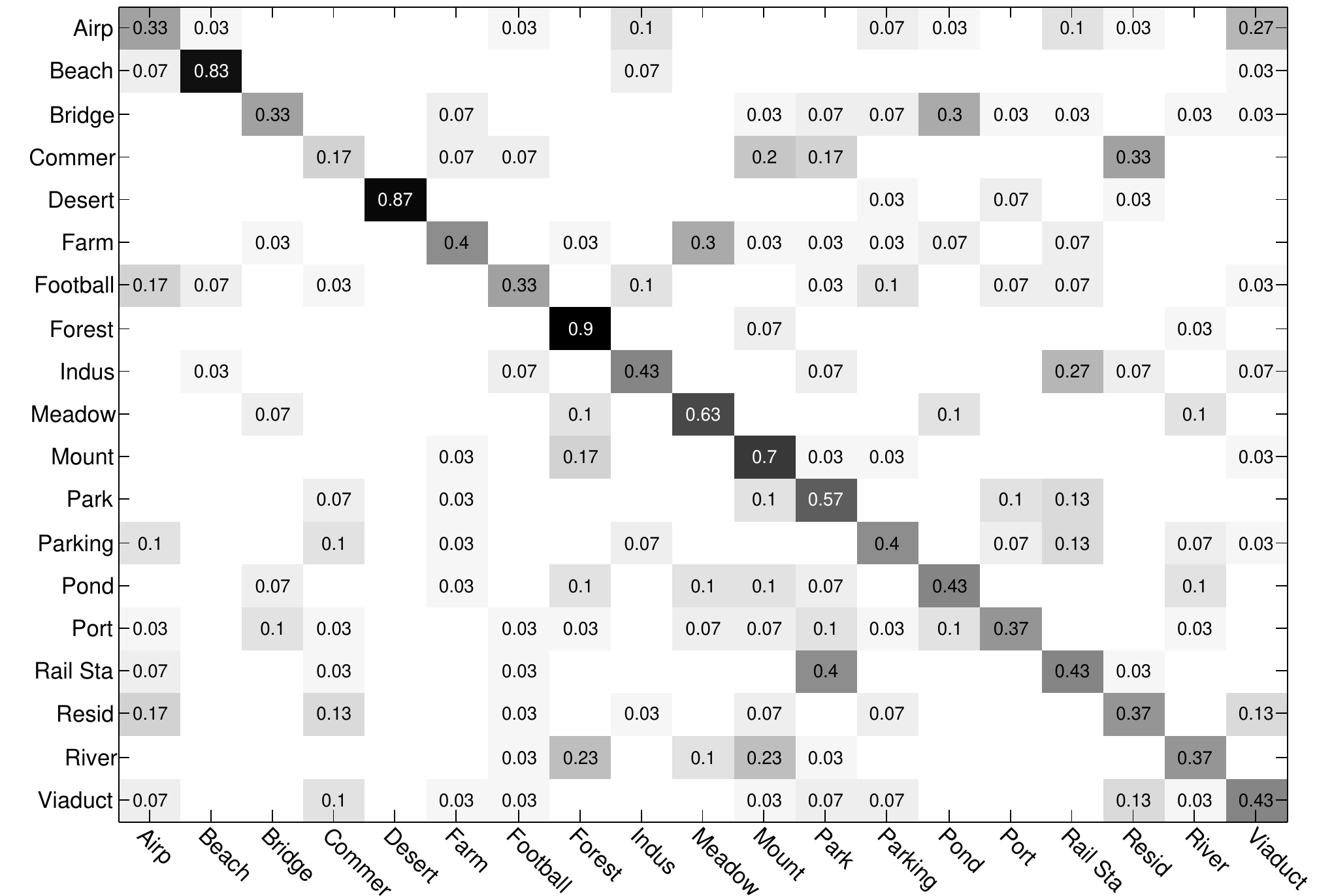}}
 \centerline{(a) low-level (CH)}
\end{minipage}
\hfill
\begin{minipage}[t]{0.33\linewidth}
 \centerline{\includegraphics[width= \linewidth]{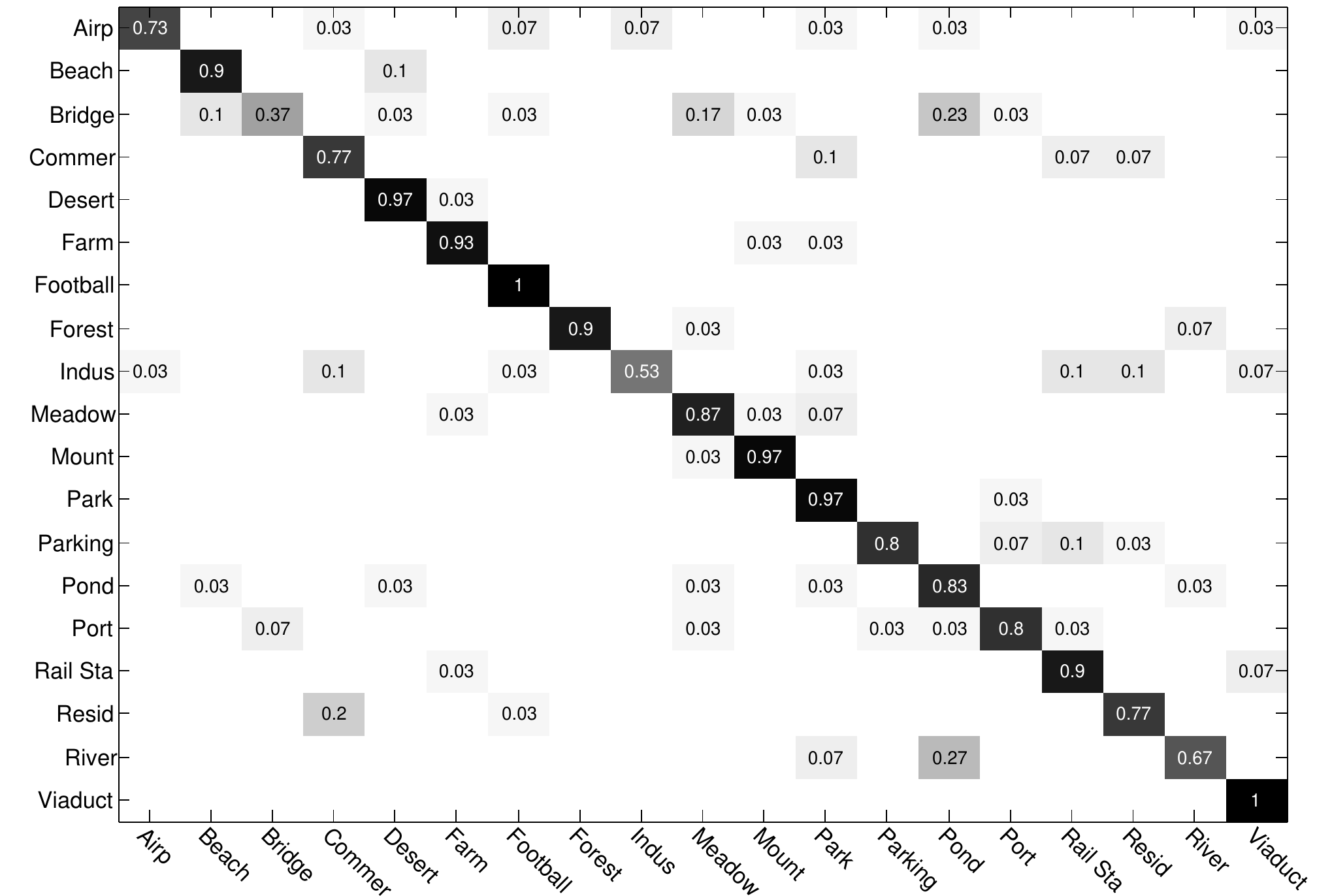}}
 \centerline{(b) mid-level IFK (SIFT)}
\end{minipage}
\hfill
\begin{minipage}[t]{0.33\linewidth}
 \centerline{\includegraphics[width=  \linewidth]{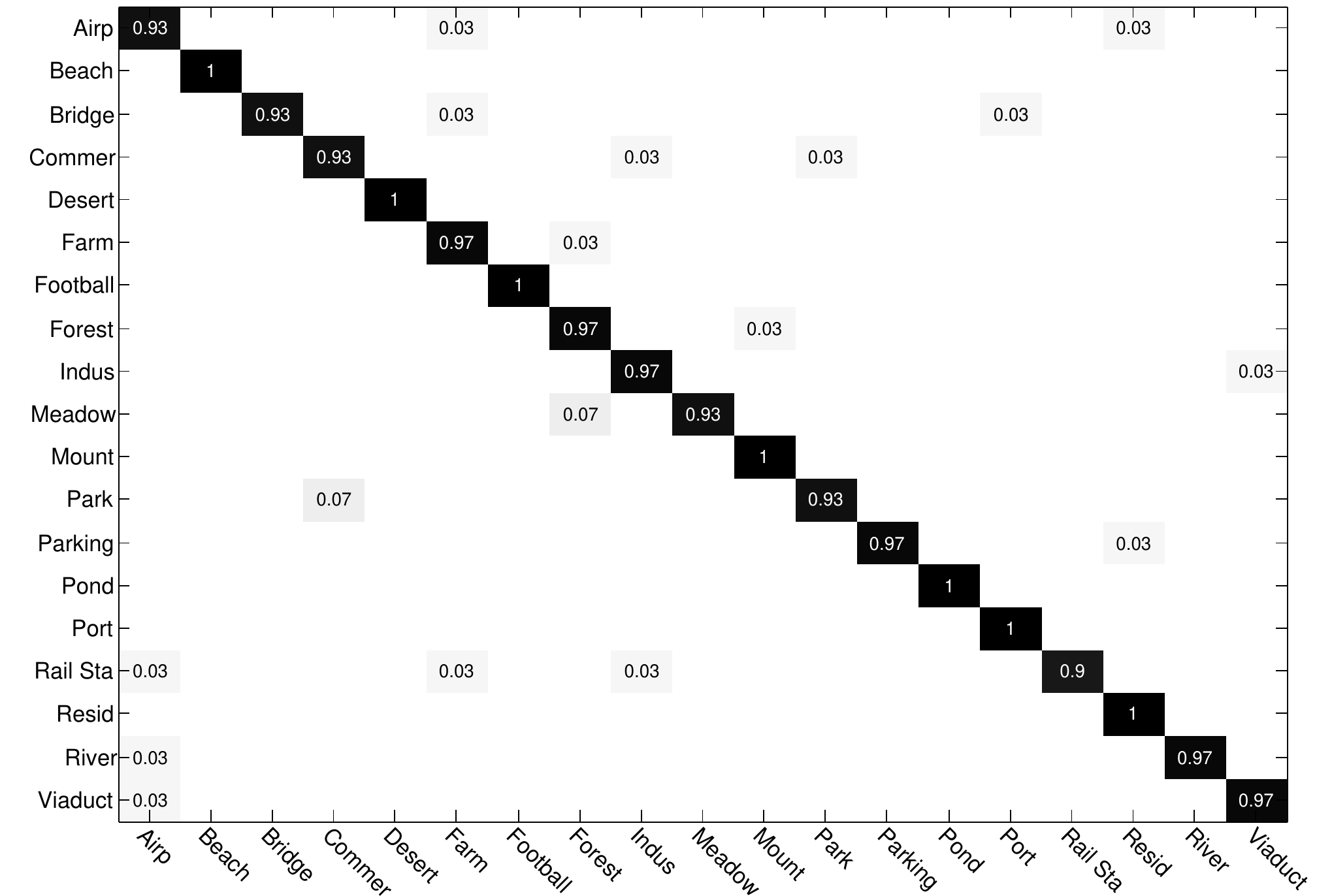}}
 \centerline{(c) high-level (VGG-VD-16)}
\end{minipage}
\end{tabular}
\caption{Confusion matrix obtained by
  low-level (CH), mid-level IFK (SIFT) and high-level (VGG-VD-16) on the WHU-RS19 dataset.}
\label{rs19-cm}
\end{figure*}

\begin{figure*}[htb!]
\centering
\begin{tabular}{ccc}
\begin{minipage}[t]{0.33\linewidth}
 \centerline{\includegraphics[width=  \linewidth]{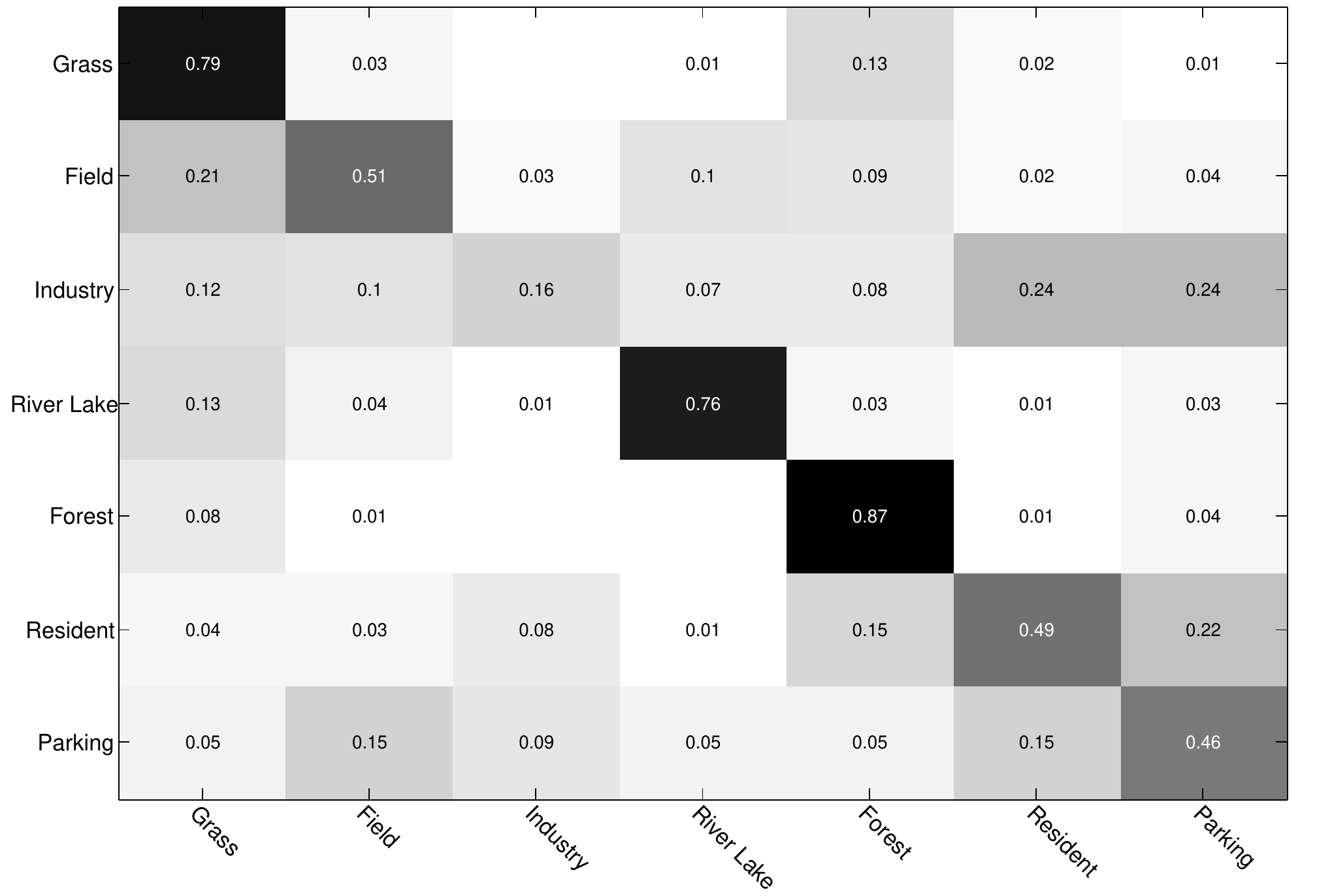}}
 \centerline{(a) low-level (LBP)}
\end{minipage}
\hfill
\begin{minipage}[t]{0.33\linewidth}
 \centerline{\includegraphics[width= \linewidth]{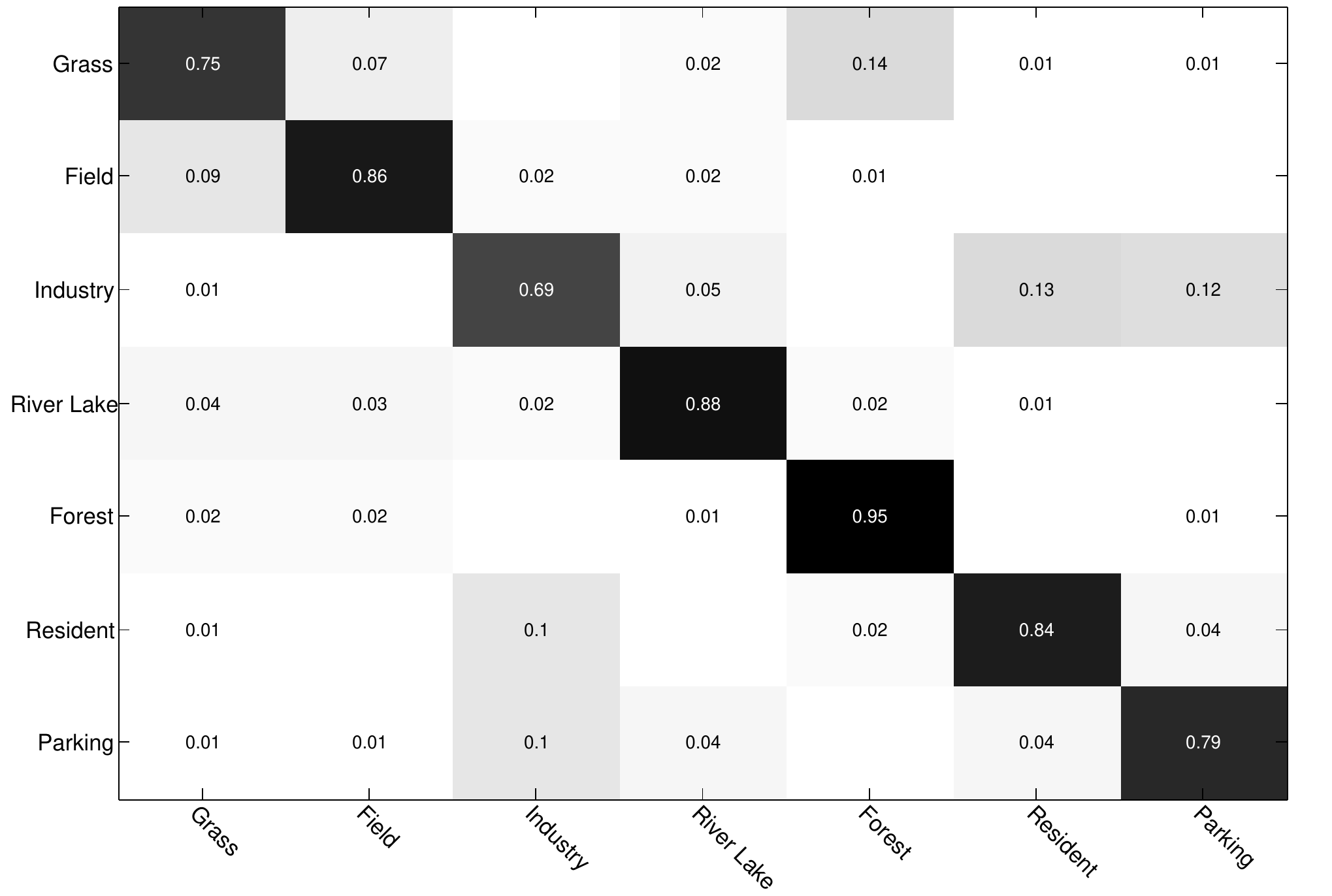}}
 \centerline{(b) mid-level IFK (SIFT)}
\end{minipage}
\hfill
\begin{minipage}[t]{0.33\linewidth}
 \centerline{\includegraphics[width=  \linewidth]{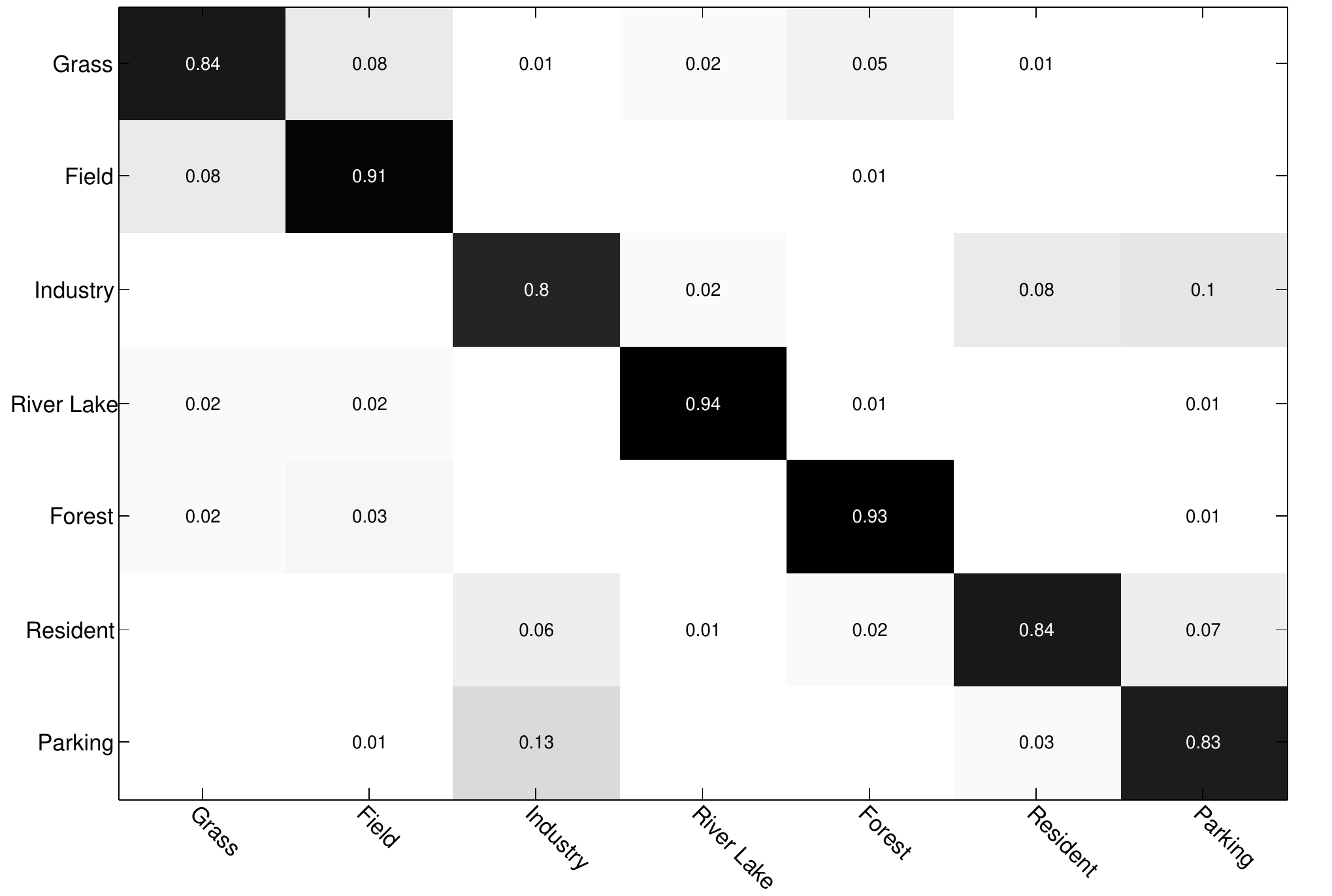}}
 \centerline{(c) high-level (CaffeNet)}
\end{minipage}
\end{tabular}
\caption{Confusion matrix obtained by
  low-level (LBP), mid-level IFK (SIFT) and high-level (CaffeNet) on the RSSCN7 dataset.}
\label{rsscn7-cm}
\end{figure*}

\begin{figure*}[htb!]
\centering
\begin{tabular}{ccc}
\begin{minipage}[t]{0.33\linewidth}
 \centerline{\includegraphics[width=  \linewidth]{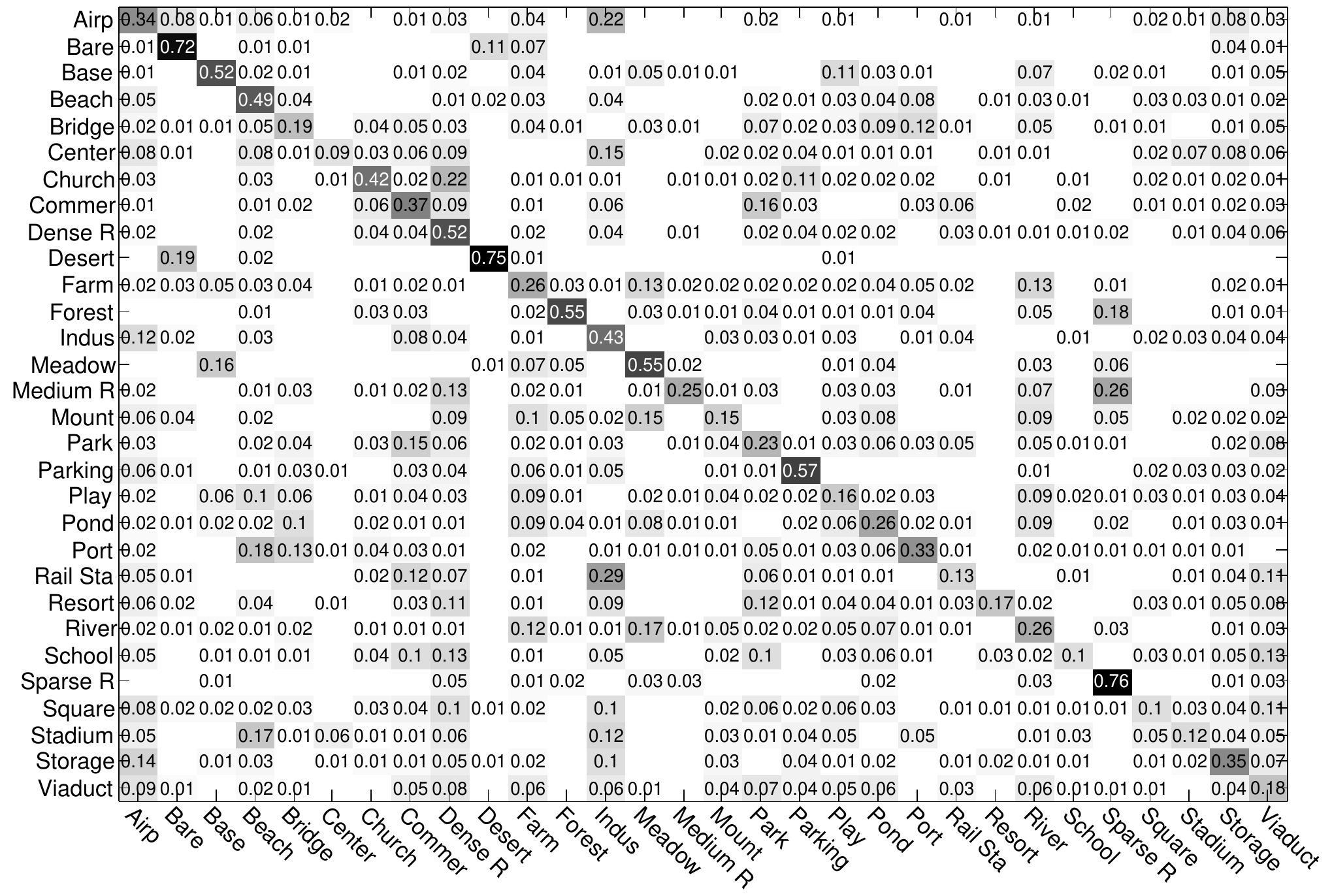}}
 \centerline{(a) low-level (CH)}
\end{minipage}
\hfill
\begin{minipage}[t]{0.33\linewidth}
 \centerline{\includegraphics[width= \linewidth]{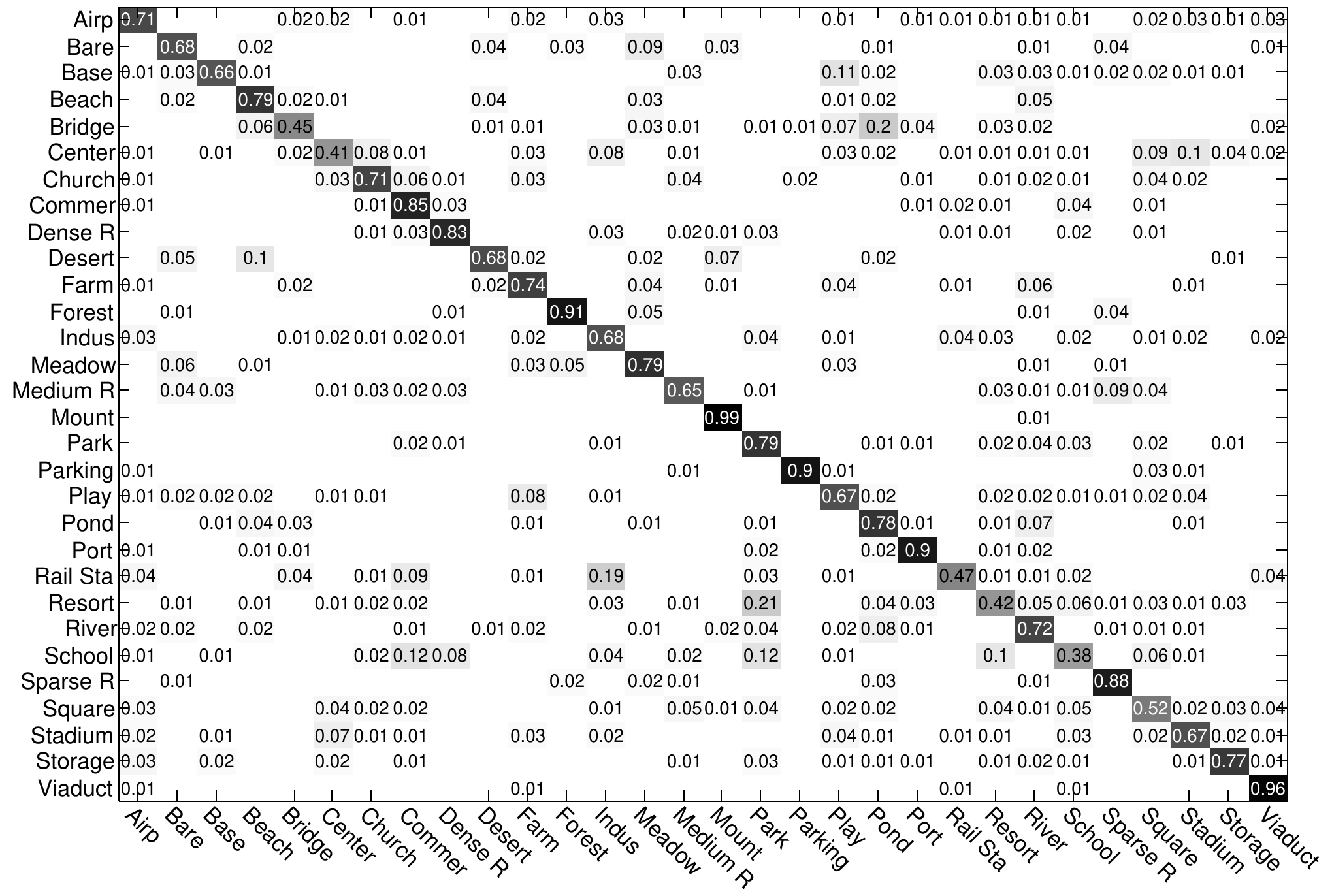}}
 \centerline{(b) mid-level IFK (SIFT)}
\end{minipage}
\hfill
\begin{minipage}[t]{0.33\linewidth}
 \centerline{\includegraphics[width=  \linewidth]{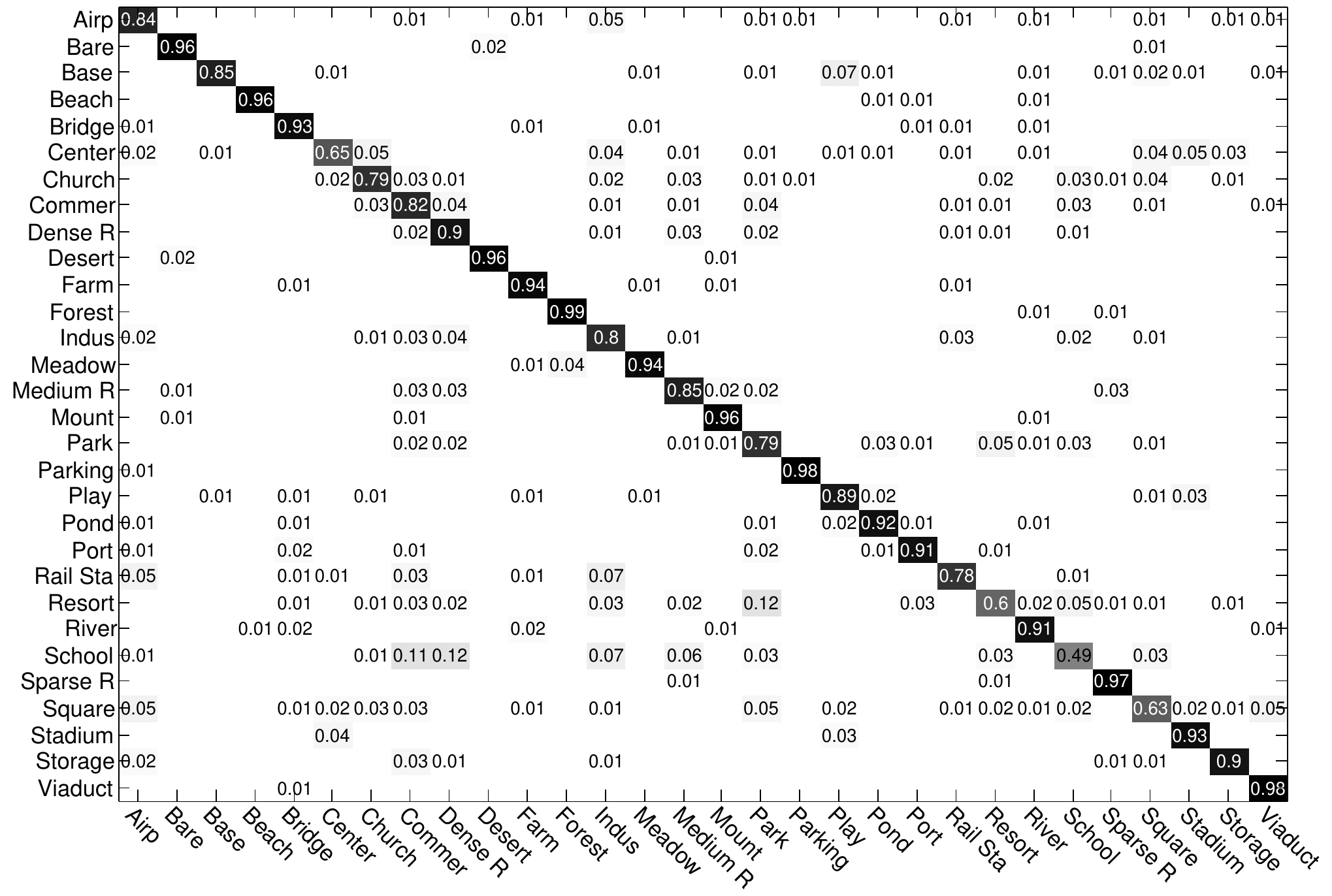}}
 \centerline{(c) high-level (CaffeNet)}
\end{minipage}
\end{tabular}
\caption{Confusion matrix obtained by
  low-level (CH), mid-level IFK (SIFT) and high-level (CaffeNet) on our AID dataset.}
\label{new-cm}
\end{figure*}

Besides giving the OA of various methods, we also compute the corresponding confusion matrix. For each dataset, we choose to show the best results of the low-level, mid-level and high-level methods for each dataset.
Fig.~\ref{ucm-cm} shows the confusion matrix using low-level (GIST), mid-level IFK (SIFT) and high-level (VGG-VD-16) on the UC-Merced dataset, and Fig.~\ref{rs19-cm} gives the results on WHU-RS19 dataset, and Fig.~\ref{rsscn7-cm} gives the results on RSSCN7 dataset, and Fig.~\ref{new-cm} is our AID dataset.

From the confusion matrix on the UC-Merced dataset (Fig.~\ref{ucm-cm}), we can see that there are only 2 classes obtain the classification accuracy above 0.8, and most classes are easily confused with others using low-level features; when using mid-level features, the classification accuracies of all the scene types increase and more than a half of the classes achieve the classification accuracy above 0.8; while for high-level features, the scene types can be easily distinguished from others that the classification accuracies of most classes are close to or even equal to 1. The most notable confusion is among buildings, dense residential, medium residential and sparse residential, for their similar structures and land cover types.

From the confusion matrix on the WHU-RS19 dataset (Fig.~\ref{rs19-cm}) and RSSCN7 dataset (Fig.~\ref{rsscn7-cm}), we can observe the similar phenomenon that low-level features can not distinguish the scene types well, and mid-level features can improve the classification accuracy a lot, while high-level features obtain a quite clean confusion matrix.

When analyzing the confusion matrix on our AID dataset (Fig.~\ref{new-cm}), similarly, the low-level features give the worst performance while high-level give the best. Although most scene types can achieve the classification accuracy close to 1 using high-level features, most of them are natural scene types and thus easy to be distinguished, e.g., bareland, beach, desert, forest, mountain , etc.. Note that the most difficult case in the UC-Merced dataset, i.e., sparse residential, medium residential and dense residential areas, the classification accuracies in our new dataset are around 0.9, which no longer belongs to the difficult ones. The most difficult scene types in our new dataset are almost newly added scene types, i.e., school (0.49), resort(0.6), square (0.63) and center(0.65). The most notable confusion is between resort and park, which may be caused by the fact that some parks are built for leisure, which have the similar appearances with resorts for holiday, especially in China, they may both contain green belts, lakes, etc., and thus are easily confused. In addition, school and dense residential are also highly confused for the densely distributed buildings.

By comparing the confusion matrix among the four datasets, we can conclude that our new dataset is more suitable for aerial scene classification than the others for it contains much fine-grained meanwhile challenging scene types.

\subsection{Discussion}

From the above experimental results, we can summarize some interesting but meaningful observations as follows:
\begin{itemize}
  \item[-] By comparing various scene classification methods, we can observe the layered performances as the name implies: low-level methods have relatively worse performances, while high-level methods perform better on all the datasets, which shows the great potential of high-level methods.
  \item[-] By comparing different scene classification datasets, we can find that our new dataset is far more challenging than the others for it has relatively higher intra-class variations and smaller inter-class dissimilarity. In addition, the large numbers of sample images can help to evaluate various methods more precisely.
\end{itemize}

The above observations can provide us with very meaningful instructions for investigating more effective high-level methods on more challenging dataset to promote the progress in aerial scene classification.

\section{Conclusion}
\label{sec:conclusion}
In this paper, we firstly give a comprehensive review on aerial scene classification by giving a clear summary of the existing approaches. We find that the results on the current popular-used datasets are already saturated and thus severely limit the progress of aerial scene classification. In order to solve the problem, we construct a new large-scale dataset, i.e. AID, which is the largest and most challenging one for the scene classification of aerial images. The purpose of the dataset is to provide the research community with a benchmark resource to advance the state-of-the-art algorithms in aerial scene analysis. In addition,
we have evaluated a set of representative aerial scene classification approaches with various experimental protocols on
the new dataset. These can serve as baseline results for
future works. Moreover, both the dataset and the codes are public online for freely downloading to promote the development of aerial scene classification.

%To solve the current situation that the lack of public challenging remote sensing datasets makes it difficult to benchmark different scene classification methods fairly, in this paper, we firstly give a comprehensive review of aerial scene classification methods, and then present a new public and large scale remote sensing dataset for fairly comparing various scene classification methods. Our benchmark dataset consists of 10000 images manually labeled into 30 classes, which is the largest and most challenging one. We describe the various properties of the new dataset, and shows the comparative results for some representative scene classification algorithms on our new dataset as well as the common used UC-Merced dataset and WHU-RS19 dataset. From the comparative results, we can get some meaningful instructions

\section*{Acknowledgment}
The authors would like to thank all the researchers who kindly sharing the codes used in our studies and all the volunteers who help us constructing the dataset. This research is supported by the National Natural Science Foundation of China under the contracts No.91338113 and No.41501462.
\bibliographystyle{IEEEtran}
\bibliography{benchmark-ref}

\begin{thebibliography}{10}
\providecommand{\url}[1]{#1}
\csname url@samestyle\endcsname
\providecommand{\newblock}{\relax}
\providecommand{\bibinfo}[2]{#2}
\providecommand{\BIBentrySTDinterwordspacing}{\spaceskip=0pt\relax}
\providecommand{\BIBentryALTinterwordstretchfactor}{4}
\providecommand{\BIBentryALTinterwordspacing}{\spaceskip=\fontdimen2\font plus
\BIBentryALTinterwordstretchfactor\fontdimen3\font minus
  \fontdimen4\font\relax}
\providecommand{\BIBforeignlanguage}[2]{{%
\expandafter\ifx\csname l@#1\endcsname\relax
\typeout{** WARNING: IEEEtran.bst: No hyphenation pattern has been}%
\typeout{** loaded for the language `#1'. Using the pattern for}%
\typeout{** the default language instead.}%
\else
\language=\csname l@#1\endcsname
\fi
#2}}
\providecommand{\BIBdecl}{\relax}
\BIBdecl

\bibitem{hu2013exploring}
Q.~Hu, W.~Wu, T.~Xia, Q.~Yu, P.~Yang, Z.~Li, and Q.~Song, ``Exploring the use
  of google earth imagery and object-based methods in land use/cover mapping,''
  \emph{Remote Sensing}, vol.~5, no.~11, pp. 6026--6042, 2013.

\bibitem{cheng2015effective}
G.~Cheng, J.~Han, L.~Guo, Z.~Liu, S.~Bu, and J.~Ren, ``Effective and efficient
  midlevel visual elements-oriented land-use classification using vhr remote
  sensing images,'' \emph{IEEE Transactions on Geoscience and Remote Sensing},
  vol.~53, no.~8, pp. 4238--4249, 2015.

\bibitem{cheng2014multi}
G.~Cheng, J.~Han, P.~Zhou, and L.~Guo, ``Multi-class geospatial object
  detection and geographic image classification based on collection of part
  detectors,'' \emph{ISPRS Journal of Photogrammetry and Remote Sensing},
  vol.~98, pp. 119--132, 2014.

\bibitem{hu2015transferring}
F.~Hu, G.-S. Xia, J.~Hu, and L.~Zhang, ``Transferring deep convolutional neural
  networks for the scene classification of high-resolution remote sensing
  imagery,'' \emph{Remote Sensing}, vol.~7, no.~11, pp. 14\,680--14\,707, 2015.

\bibitem{risojevic2011aerial}
V.~Risojevi{\'c} and Z.~Babi{\'c}, ``Aerial image classification using
  structural texture similarity,'' in \emph{IEEE International Symposium on
  Signal Processing and Information Technology (ISSPIT)}.\hskip 1em plus 0.5em
  minus 0.4em\relax IEEE, 2011, pp. 190--195.

\bibitem{yang2011spatial}
Y.~Yang and S.~Newsam, ``Spatial pyramid co-occurrence for image
  classification,'' in \emph{IEEE International Conference on Computer Vision
  (ICCV)}.\hskip 1em plus 0.5em minus 0.4em\relax IEEE, 2011, pp. 1465--1472.

\bibitem{sheng2012high}
G.~Sheng, W.~Yang, T.~Xu, and H.~Sun, ``High-resolution satellite scene
  classification using a sparse coding based multiple feature combination,''
  \emph{International journal of remote sensing}, vol.~33, no.~8, pp.
  2395--2412, 2012.

\bibitem{risojevic2012orientation}
V.~Risojevi{\'c} and Z.~Babi{\'c}, ``Orientation difference descriptor for
  aerial image classification,'' in \emph{International Conference on Systems,
  Signals and Image Processing (IWSSIP)}.\hskip 1em plus 0.5em minus
  0.4em\relax IEEE, 2012, pp. 150--153.

\bibitem{hu2013tile}
F.~Hu, W.~Yang, J.~Chen, and H.~Sun, ``Tile-level annotation of satellite
  images using multi-level max-margin discriminative random field,''
  \emph{Remote Sensing}, vol.~5, no.~5, pp. 2275--2291, 2013.

\bibitem{luo2013indexing}
B.~Luo, S.~Jiang, and L.~Zhang, ``Indexing of remote sensing images with
  different resolutions by multiple features,'' \emph{IEEE Journal of Selected
  Topics in Applied Earth Observations and Remote Sensing}, vol.~6, no.~4, pp.
  1899--1912, 2013.

\bibitem{shao2013hierarchical}
W.~Shao, W.~Yang, G.-S. Xia, and G.~Liu, ``A hierarchical scheme of multiple
  feature fusion for high-resolution satellite scene categorization,'' in
  \emph{Computer Vision Systems}.\hskip 1em plus 0.5em minus 0.4em\relax
  Springer, 2013, pp. 324--333.

\bibitem{shao2013extreme}
W.~Shao, W.~Yang, and G.-S. Xia, ``Extreme value theory-based calibration for
  the fusion of multiple features in high-resolution satellite scene
  classification,'' \emph{International Journal of Remote Sensing}, vol.~34,
  no.~23, pp. 8588--8602, 2013.

\bibitem{risojevic2013fusion}
V.~Risojevic and Z.~Babic, ``Fusion of global and local descriptors for remote
  sensing image classification,'' \emph{IEEE Geoscience and Remote Sensing
  Letters}, vol.~10, no.~4, pp. 836--840, 2013.

\bibitem{yang2013geographic}
Y.~Yang and S.~Newsam, ``Geographic image retrieval using local invariant
  features,'' \emph{IEEE Transactions on Geoscience and Remote Sensing},
  vol.~51, no.~2, pp. 818--832, 2013.

\bibitem{zhao2013scene}
B.~Zhao, Y.~Zhong, and L.~Zhang, ``Scene classification via latent dirichlet
  allocation using a hybrid generative/discriminative strategy for high spatial
  resolution remote sensing imagery,'' \emph{Remote Sensing Letters}, vol.~4,
  no.~12, pp. 1204--1213, 2013.

\bibitem{zhao2013hybrid}
------, ``Hybrid generative/discriminative scene classification strategy based
  on latent dirichlet allocation for high spatial resolution remote sensing
  imagery,'' in \emph{IEEE International Geoscience and Remote Sensing
  Symposium (IGARSS)}.\hskip 1em plus 0.5em minus 0.4em\relax IEEE, 2013, pp.
  196--199.

\bibitem{zheng2013automatic}
X.~Zheng, X.~Sun, K.~Fu, and H.~Wang, ``Automatic annotation of satellite
  images via multifeature joint sparse coding with spatial relation
  constraint,'' \emph{IEEE Geoscience and Remote Sensing Letters}, vol.~10,
  no.~4, pp. 652--656, 2013.

\bibitem{avramovic2014block}
A.~Avramovi{\'c} and V.~Risojevi{\'c}, ``Block-based semantic classification of
  high-resolution multispectral aerial images,'' \emph{Signal, Image and Video
  Processing}, pp. 1--10, 2014.

\bibitem{cheriyadat2014unsupervised}
A.~M. Cheriyadat, ``Unsupervised feature learning for aerial scene
  classification,'' \emph{IEEE Transactions on Geoscience and Remote Sensing},
  vol.~52, no.~1, pp. 439--451, 2014.

\bibitem{kusumaningrum2014integrated}
R.~Kusumaningrum, H.~Wei, R.~Manurung, and A.~Murni, ``Integrated visual
  vocabulary in latent dirichlet allocation--based scene classification for
  ikonos image,'' \emph{Journal of Applied Remote Sensing}, vol.~8, no.~1, pp.
  083\,690--083\,690, 2014.

\bibitem{negrel2014evaluation}
R.~Negrel, D.~Picard, and P.-H. Gosselin, ``Evaluation of second-order visual
  features for land-use classification,'' in \emph{International Workshop on
  Content-Based Multimedia Indexing (CBMI)}.\hskip 1em plus 0.5em minus
  0.4em\relax IEEE, 2014, pp. 1--5.

\bibitem{zhao2014wavelet}
L.~Zhao, P.~Tang, and L.~Huo, ``A 2-d wavelet decomposition-based
  bag-of-visual-words model for land-use scene classification,''
  \emph{International Journal of Remote Sensing}, vol.~35, no.~6, pp.
  2296--2310, 2014.

\bibitem{zhao2014land}
L.-J. Zhao, P.~Tang, and L.-Z. Huo, ``Land-use scene classification using a
  concentric circle-structured multiscale bag-of-visual-words model,''
  \emph{IEEE Journal of Selected Topics in Applied Earth Observations and
  Remote Sensing}, vol.~7, no.~12, pp. 4620--4631, 2014.

\bibitem{zhu2014multi}
Q.~Zhu, Y.~Zhong, and L.~Zhang, ``Multi-feature probability topic scene
  classifier for high spatial resolution remote sensing imagery,'' in
  \emph{IEEE International Geoscience and Remote Sensing Symposium
  (IGARSS)}.\hskip 1em plus 0.5em minus 0.4em\relax IEEE, 2014, pp. 2854--2857.

\bibitem{chen2015pyramid}
S.~Chen and Y.~Tian, ``Pyramid of spatial relatons for scene-level land use
  classification,'' \emph{IEEE Transactions on Geoscience and Remote Sensing},
  vol.~53, no.~4, pp. 1947--1957, 2015.

\bibitem{chen2015measuring}
X.~Chen, T.~Fang, H.~Huo, and D.~Li, ``Measuring the effectiveness of various
  features for thematic information extraction from very high resolution remote
  sensing imagery,'' \emph{IEEE Transactions on Geoscience and Remote Sensing},
  vol.~53, no.~9, pp. 4837--4851, 2015.

\bibitem{zhong2015scene}
Y.~Zhong, M.~Cui, Q.~Zhu, and L.~Zhang, ``Scene classification based on
  multifeature probabilistic latent semantic analysis for high spatial
  resolution remote sensing images,'' \emph{Journal of Applied Remote Sensing},
  vol.~9, no.~1, pp. 095\,064--095\,064, 2015.

\bibitem{sridharan2015bag}
H.~Sridharan and A.~Cheriyadat, ``Bag of lines (bol) for improved aerial scene
  representation,'' \emph{IEEE Geoscience and Remote Sensing Letters}, vol.~12,
  no.~3, pp. 676--680, 2015.

\bibitem{hu2015comparative}
J.~Hu, G.-S. Xia, F.~Hu, and L.~Zhang, ``A comparative study of sampling
  analysis in the scene classification of optical high-spatial resolution
  remote sensing imagery,'' \emph{Remote Sensing}, vol.~7, no.~11, pp.
  14\,988--15\,013, 2015.

\bibitem{hu2015benchmark}
J.~Hu, T.~Jiang, X.~Tong, G.-S. Xia, and L.~Zhang, ``A benchmark for scene
  classification of high spatial resolution remote sensing imagery,'' in
  \emph{IEEE International Geoscience and Remote Sensing Symposium
  (IGARSS)}.\hskip 1em plus 0.5em minus 0.4em\relax IEEE, 2015, pp. 5003--5006.

\bibitem{hu2015unsupervised}
F.~Hu, G.-S. Xia, Z.~Wang, X.~Huang, L.~Zhang, and H.~Sun, ``Unsupervised
  feature learning via spectral clustering of multidimensional patches for
  remotely sensed scene classification,'' \emph{IEEE Journal of Selected Topics
  in Applied Earth Observations and Remote Sensing}, vol.~8, no.~5, pp.
  2015--2030, 2015.

\bibitem{castelluccio2015land}
M.~Castelluccio, G.~Poggi, C.~Sansone, and L.~Verdoliva, ``Land use
  classification in remote sensing images by convolutional neural networks,''
  \emph{arXiv preprint arXiv:1508.00092}, 2015.

\bibitem{penatti2015deep}
O.~A.~B. Penatti, K.~Nogueira, and J.~A. dos Santos, ``Do deep features
  generalize from everyday objects to remote sensing and aerial scenes
  domains?'' in \emph{Proc. IEEE Conference on Computer Vision and Pattern
  Recognition}, June 2015.

\bibitem{luus2015multiview}
F.~Luus, B.~Salmon, F.~van~den Bergh, and B.~Maharaj, ``Multiview deep learning
  for land-use classification,'' \emph{IEEE Geoscience and Remote Sensing
  Letters}, vol.~12, no.~12, pp. 2448--2452, 2015.

\bibitem{yang2015learning}
W.~Yang, X.~Yin, and G.-S. Xia, ``Learning high-level features for satellite
  image classification with limited labeled samples,'' \emph{IEEE Transactions
  on Geoscience and Remote Sensing}, vol.~53, no.~8, pp. 4472--4482, 2015.

\bibitem{zhang2015saliency}
F.~Zhang, B.~Du, and L.~Zhang, ``Saliency-guided unsupervised feature learning
  for scene classification,'' \emph{IEEE Transactions on Geoscience and Remote
  Sensing}, vol.~53, no.~4, pp. 2175--2184, 2015.

\bibitem{zhang2015scene}
------, ``Scene classification via a gradient boosting random convolutional
  network framework,'' \emph{IEEE Transactions on Geoscience and Remote
  Sensing}, vol.~PP, no.~99, pp. 1--10, 2015.

\bibitem{zhu2015scene}
Y.~Zhong, Q.~Zhu, and L.~Zhang, ``Scene classification based on the
  multifeature fusion probabilistic topic model for high spatial resolution
  remote sensing imagery,'' \emph{IEEE Transactions on Geoscience and Remote
  Sensing}, vol.~53, no.~11, pp. 6207--6222, 2015.

\bibitem{chen2015land}
C.~Chen, B.~Zhang, H.~Su, W.~Li, and L.~Wang, ``Land-use scene classification
  using multi-scale completed local binary patterns,'' \emph{Signal, Image and
  Video Processing}, pp. 1--8, 2015.

\bibitem{zou2015deep}
Q.~Zou, L.~Ni, T.~Zhang, and Q.~Wang, ``Deep learning based feature selection
  for remote sensing scene classification,'' \emph{Geoscience and Remote
  Sensing Letters, IEEE}, vol.~12, no.~11, pp. 2321--2325, 2015.

\bibitem{nogueira2016towards}
K.~Nogueira, O.~A. Penatti, and J.~A.~d. Santos, ``Towards better exploiting
  convolutional neural networks for remote sensing scene classification,''
  \emph{arXiv preprint arXiv:1602.01517}, 2016.

\bibitem{tuia2009active}
D.~Tuia, F.~Ratle, F.~Pacifici, M.~F. Kanevski, and W.~J. Emery, ``Active
  learning methods for remote sensing image classification,'' \emph{IEEE
  Transactions on Geoscience and Remote Sensing}, vol.~47, no.~7, pp.
  2218--2232, 2009.

\bibitem{tuia2011survey}
D.~Tuia, M.~Volpi, L.~Copa, M.~Kanevski, and J.~Mu{\~n}oz-Mar{\'\i}, ``A survey
  of active learning algorithms for supervised remote sensing image
  classification,'' \emph{IEEE Journal of Selected Topics in Signal
  Processing}, vol.~5, no.~3, pp. 606--617, 2011.

\bibitem{blaschke2001pixels}
T.~Blaschke and J.~Strobl, ``What¡¯s wrong with pixels? some recent
  developments interfacing remote sensing and gis,'' \emph{GeoBIT/GIS}, vol.~6,
  no.~01, pp. 12--17, 2001.

\bibitem{kotliar1990multiple}
N.~B. Kotliar and J.~A. Wiens, ``Multiple scales of patchiness and patch
  structure: a hierarchical framework for the study of heterogeneity,''
  \emph{Oikos}, pp. 253--260, 1990.

\bibitem{blaschke2003object}
T.~Blaschke, ``Object-based contextual image classification built on image
  segmentation,'' in \emph{IEEE Workshop on Advances in Techniques for Analysis
  of Remotely Sensed Data}.\hskip 1em plus 0.5em minus 0.4em\relax IEEE, 2003,
  pp. 113--119.

\bibitem{yan2006comparison}
G.~Yan, J.-F. Mas, B.~Maathuis, Z.~Xiangmin, and P.~Van~Dijk, ``Comparison of
  pixel-based and object-oriented image classification approaches¡ªa case study
  in a coal fire area, wuda, inner mongolia, china,'' \emph{International
  Journal of Remote Sensing}, vol.~27, no.~18, pp. 4039--4055, 2006.

\bibitem{blaschke2010object}
T.~Blaschke, ``Object based image analysis for remote sensing,'' \emph{ISPRS
  journal of photogrammetry and remote sensing}, vol.~65, no.~1, pp. 2--16,
  2010.

\bibitem{myint2011per}
S.~W. Myint, P.~Gober, A.~Brazel, S.~Grossman-Clarke, and Q.~Weng, ``Per-pixel
  vs. object-based classification of urban land cover extraction using high
  spatial resolution imagery,'' \emph{Remote sensing of environment}, vol. 115,
  no.~5, pp. 1145--1161, 2011.

\bibitem{duro2012comparison}
D.~C. Duro, S.~E. Franklin, and M.~G. Dub{\'e}, ``A comparison of pixel-based
  and object-based image analysis with selected machine learning algorithms for
  the classification of agricultural landscapes using spot-5 hrg imagery,''
  \emph{Remote Sensing of Environment}, vol. 118, pp. 259--272, 2012.

\bibitem{zhong2014hybrid}
Y.~Zhong, J.~Zhao, and L.~Zhang, ``A hybrid object-oriented conditional random
  field classification framework for high spatial resolution remote sensing
  imagery,'' \emph{IEEE Transactions on Geoscience and Remote Sensing},
  vol.~52, no.~11, pp. 7023--7037, 2014.

\bibitem{zhao2015detail}
J.~Zhao, Y.~Zhong, and L.~Zhang, ``Detail-preserving smoothing classifier based
  on conditional random fields for high spatial resolution remote sensing
  imagery,'' \emph{IEEE Transactions on Geoscience and Remote Sensing},
  vol.~53, no.~5, pp. 2440--2452, 2015.

\bibitem{yang2008comparing}
Y.~Yang and S.~Newsam, ``Comparing sift descriptors and gabor texture features
  for classification of remote sensed imagery,'' in \emph{IEEE International
  Conference on Image Processing}.\hskip 1em plus 0.5em minus 0.4em\relax IEEE,
  2008, pp. 1852--1855.

\bibitem{dos2010evaluating}
J.~A. dos Santos, O.~A.~B. Penatti, and R.~da~Silva~Torres, ``Evaluating the
  potential of texture and color descriptors for remote sensing image retrieval
  and classification.'' in \emph{VISAPP (2)}, 2010, pp. 203--208.

\bibitem{lienou2010semantic}
M.~Li{\'e}nou, H.~Ma{\^\i}tre, and M.~Datcu, ``Semantic annotation of satellite
  images using latent dirichlet allocation,'' \emph{IEEE Geoscience and Remote
  Sensing Letters}, vol.~7, no.~1, pp. 28--32, 2010.

\bibitem{xia2010structural}
G.-S. Xia, W.~Yang, J.~Delon, Y.~Gousseau, H.~Sun, and H.~Ma{\^\i}tre,
  ``Structural high-resolution satellite image indexing,'' in \emph{ISPRS TC
  VII Symposium-100 Years ISPRS}, vol.~38, 2010, pp. 298--303.

\bibitem{yang2010bag}
Y.~Yang and S.~Newsam, ``Bag-of-visual-words and spatial extensions for
  land-use classification,'' in \emph{Proceedings of the 18th SIGSPATIAL
  International Conference on Advances in Geographic Information
  Systems}.\hskip 1em plus 0.5em minus 0.4em\relax ACM, 2010, pp. 270--279.

\bibitem{chen2011evaluation}
L.~Chen, W.~Yang, K.~Xu, and T.~Xu, ``Evaluation of local features for scene
  classification using vhr satellite images,'' in \emph{Joint Urban Remote
  Sensing Event (JURSE)}.\hskip 1em plus 0.5em minus 0.4em\relax IEEE, 2011,
  pp. 385--388.

\bibitem{dai2011satellite}
D.~Dai and W.~Yang, ``Satellite image classification via two-layer sparse
  coding with biased image representation,'' \emph{IEEE Geoscience and Remote
  Sensing Letters}, vol.~8, no.~1, pp. 173--176, 2011.

\bibitem{risojevic2011gabor}
V.~Risojevi{\'c}, S.~Momi{\'c}, and Z.~Babi{\'c}, ``Gabor descriptors for
  aerial image classification,'' in \emph{Adaptive and Natural Computing
  Algorithms}.\hskip 1em plus 0.5em minus 0.4em\relax Springer, 2011, pp.
  51--60.

\bibitem{oliva2001modeling}
A.~Oliva and A.~Torralba, ``Modeling the shape of the scene: A holistic
  representation of the spatial envelope,'' \emph{International Journal of
  Computer Vision}, vol.~42, no.~3, pp. 145--175, 2001.

\bibitem{lowe2004sift}
D.~G. Lowe, ``Distinctive image features from scale-invariant keypoints,''
  \emph{International Journal of Computer Vision}, vol.~60, no.~2, pp. 91--110,
  2004.

\bibitem{swain1991color}
M.~J. Swain and D.~H. Ballard, ``Color indexing,'' \emph{International journal
  of computer vision}, vol.~7, no.~1, pp. 11--32, 1991.

\bibitem{manjunath1996texture}
B.~S. Manjunath and W.-Y. Ma, ``Texture features for browsing and retrieval of
  image data,'' \emph{IEEE Transactions on Pattern Analysis and Machine
  Intelligence}, vol.~18, no.~8, pp. 837--842, 1996.

\bibitem{ojala2002lbp}
T.~Ojala, M.~Pietik{\"a}inen, and T.~M{\"a}enp{\"a}{\"a}, ``Multiresolution
  gray-scale and rotation invariant texture classification with local binary
  patterns,'' \emph{IEEE Transactions on Pattern Analysis and Machine
  Intelligence}, vol.~24, no.~7, pp. 971--987, 2002.

\bibitem{luo2008indexing}
B.~Luo, J.-F. Aujol, Y.~Gousseau, and S.~Ladjal, ``Indexing of satellite images
  with different resolutions by wavelet features,'' \emph{IEEE Transactions on
  Image Processing}, vol.~17, no.~8, pp. 1465--1472, 2008.

\bibitem{luo2009local}
B.~Luo, J.-F. Aujol, and Y.~Gousseau, ``Local scale measure from the
  topographic map and application to remote sensing images,'' \emph{Multiscale
  modeling \& simulation}, vol.~8, no.~1, pp. 1--29, 2009.

\bibitem{mallat2012combined}
S.~Mallat and L.~Sifre, ``Combined scattering for rotation invariant texture
  analysis,'' \emph{submitted to ESANN}, 2012.

\bibitem{scheirer2012evt}
W.~J. Scheirer, N.~Kumar, P.~N. Belhumeur, and T.~E. Boult, ``Multi-attribute
  spaces: Calibration for attribute fusion and similarity search,'' in
  \emph{IEEE Conference on Computer Vision and Pattern Recognition}.\hskip 1em
  plus 0.5em minus 0.4em\relax IEEE, 2012, pp. 2933--2940.

\bibitem{yang2009scspm}
J.~Yang, K.~Yu, Y.~Gong, and T.~Huang, ``Linear spatial pyramid matching using
  sparse coding for image classification,'' in \emph{Proc. IEEE Conference on
  Computer Vision and Pattern Recognition}, 2009, pp. 1794--1801.

\bibitem{Perronnin2010ifk}
F.~Perronnin, J.~S\'{a}nchez, and T.~Mensink, ``Improving the fisher kernel for
  large-scale image classification,'' in \emph{Proc. European Conference on
  Computer Vision}, 2010, pp. 143--156.

\bibitem{lazebnik2006spm}
S.~Lazebnik, C.~Schmid, and J.~Ponce, ``Beyond bags of features: Spatial
  pyramid matching for recognizing natural scene categories,'' in \emph{Proc.
  IEEE Conference on Computer Vision and Pattern Recognition}, vol.~2, 2006,
  pp. 2169--2178.

\bibitem{blei2003lda}
D.~M. Blei, A.~Y. Ng, and M.~I. Jordan, ``Latent dirichlet allocation,''
  \emph{the Journal of Machine Learning research}, vol.~3, pp. 993--1022, 2003.

\bibitem{stricker1995similarity}
M.~A. Stricker and M.~Orengo, ``Similarity of color images,'' in
  \emph{IS\&T/SPIE's Symposium on Electronic Imaging: Science \&
  Technology}.\hskip 1em plus 0.5em minus 0.4em\relax International Society for
  Optics and Photonics, 1995, pp. 381--392.

\bibitem{haralick1973textural}
R.~M. Haralick, K.~Shanmugam, and I.~H. Dinstein, ``Textural features for image
  classification,'' \emph{IEEE Transactions on Systems, Man and Cybernetics},
  no.~6, pp. 610--621, 1973.

\bibitem{bosch2006plsa}
A.~Bosch, A.~Zisserman, and X.~Mu{\~n}oz, ``Scene classification via plsa,'' in
  \emph{Proc. European Conference on Computer Vision}, 2006, pp. 517--530.

\bibitem{vincent2010ae}
P.~Vincent, H.~Larochelle, I.~Lajoie, Y.~Bengio, and P.-A. Manzagol, ``Stacked
  denoising autoencoders: Learning useful representations in a deep network
  with a local denoising criterion,'' \emph{The Journal of Machine Learning
  Research}, vol.~11, pp. 3371--3408, 2010.

\bibitem{ILSVRC15}
O.~Russakovsky, J.~Deng, H.~Su, J.~Krause, S.~Satheesh, S.~Ma, Z.~Huang,
  A.~Karpathy, A.~Khosla, M.~Bernstein, A.~C. Berg, and L.~Fei-Fei, ``{ImageNet
  Large Scale Visual Recognition Challenge},'' \emph{International Journal of
  Computer Vision}, pp. 1--42, April 2015.

\bibitem{sermanet2013overfeat}
P.~Sermanet, D.~Eigen, X.~Zhang, M.~Mathieu, R.~Fergus, and Y.~LeCun,
  ``Overfeat: Integrated recognition, localization and detection using
  convolutional networks,'' \emph{arXiv preprint arXiv:1312.6229}, 2013.

\bibitem{jia2014caffe}
Y.~Jia, E.~Shelhamer, J.~Donahue, S.~Karayev, J.~Long, R.~Girshick,
  S.~Guadarrama, and T.~Darrell, ``Caffe: Convolutional architecture for fast
  feature embedding,'' in \emph{Proceedings of the ACM International Conference
  on Multimedia}.\hskip 1em plus 0.5em minus 0.4em\relax ACM, 2014, pp.
  675--678.

\bibitem{szegedy2014going}
C.~Szegedy, W.~Liu, Y.~Jia, P.~Sermanet, S.~Reed, D.~Anguelov, D.~Erhan,
  V.~Vanhoucke, and A.~Rabinovich, ``Going deeper with convolutions,''
  \emph{arXiv preprint arXiv:1409.4842}, 2014.

\bibitem{sivic2003bow}
J.~Sivic and A.~Zisserman, ``Video google: A text retrieval approach to object
  matching in videos,'' in \emph{Proc. IEEE International Conference on
  Computer Vision}, 2003, pp. 1470--1477.

\bibitem{jegou2012vlad}
H.~J{\'e}gou, F.~Perronnin, M.~Douze, J.~Sanchez, P.~Perez, and C.~Schmid,
  ``Aggregating local image descriptors into compact codes,'' \emph{IEEE
  Transactions on Pattern Analysis and Machine Intelligence}, vol.~34, no.~9,
  pp. 1704--1716, 2012.

\bibitem{Hinton2006A}
G.~E. Hinton, S.~Osindero, and Y.~W. Teh, ``A fast learning algorithm for deep
  belief nets.'' \emph{Neural Computation}, vol.~18, no.~7, pp. 1527--54, 2006.

\bibitem{wang2010llc}
J.~Wang, J.~Yang, K.~Yu, F.~Lv, T.~Huang, and Y.~Gong, ``Locality-constrained
  linear coding for image classification,'' in \emph{Proc. IEEE Conference on
  Computer Vision and Pattern Recognition}.\hskip 1em plus 0.5em minus
  0.4em\relax IEEE, 2010, pp. 3360--3367.

\bibitem{yu2009lcc}
K.~Yu, T.~Zhang, and Y.~Gong, ``Nonlinear learning using local coordinate
  coding,'' in \emph{Advances in Neural Information Processing Systems}, 2009,
  pp. 2223--2231.

\bibitem{perronnin2007fisher}
F.~Perronnin and C.~Dance, ``Fisher kernels on visual vocabularies for image
  categorization,'' in \emph{Proc. IEEE Conference on Computer Vision and
  Pattern Recognition}.\hskip 1em plus 0.5em minus 0.4em\relax IEEE, 2007, pp.
  1--8.

\bibitem{krizhevsky2012imagenet}
A.~Krizhevsky, I.~Sutskever, and G.~E. Hinton, ``Imagenet classification with
  deep convolutional neural networks,'' in \emph{Advances in Neural Information
  Processing Systems}, 2012, pp. 1097--1105.

\bibitem{Simonyan14vggvd}
K.~Simonyan and A.~Zisserman, ``Very deep convolutional networks for
  large-scale image recognition,'' \emph{CoRR}, vol. abs/1409.1556, 2014.

\bibitem{lin2013network}
L.~Min, C.~Qiang, and S.~Yan, ``Network in network,'' \emph{CoRR}, vol.
  abs/1312.4400, 2013.

\bibitem{Fan2010LIBLINEAR}
R.~E. Fan, K.~W. Chang, C.~J. Hsieh, X.~R. Wang, and C.~J. Lin, ``Liblinear: A
  library for large linear classification,'' \emph{Journal of Machine Learning
  Research}, vol.~9, no.~12, pp. 1871--1874, 2010.

\end{thebibliography}
\end{document}